\documentclass[twoside,11pt]{article}

\usepackage[accepted]{melba}

\usepackage{amsmath,amsfonts}
\usepackage{bbm}
\usepackage{amssymb}

\usepackage{graphicx}
\usepackage{subfig}
\usepackage{float}
\usepackage[export]{adjustbox}

\usepackage{makecell}
\usepackage{multirow}
\usepackage{arydshln}

\usepackage[toc,page]{appendix}

\usepackage{xcolor}

\usepackage{ulem} % for strikethrough with \sout, adds an underline to the citations. Weird!

\newcommand{\refappendix}[1]{\textcolor{blue}{\hyperref[#1]{Appendix~\ref*{#1}}}}

\def\blueautoref#1{\textcolor{blue}{\autoref{#1}}}

\renewcommand{\sout}[1]{\makebox[0pt]{\phantom{#1}}}

\newcommand{\editedd}[1]{%
    \textcolor{black}{#1}%
}
\newcommand{\editeddd}[1]{%
    \textcolor{black}{#1}%
}

\definecolor{mycyan}{rgb}{0,1,1}
\melbaid{2024:018}  % This is provided upon by the publishing editor
\doi{https://doi.org/10.59275/j.melba.2024-5gc8}
\melbaauthors{Mody}  % Note: this one is also used to set the pdf 'authors' metadata
\volume{2}
\firstpageno{1048}  % Communicated by the publishing editor
\melbayear{2024}  % The publication year
\datesubmitted{08/2023}  % Date submitted to MELBA: mm/yyyy
\datepublished{08/2024}  % Today's date: mm/yyyy

\ShortHeadings{Improving Uncertainty-Error Correspondence}{Mody and Staring}

\title{Improving Uncertainty-Error Correspondence in Deep Bayesian Medical Image Segmentation}

\author{
    \firstname Prerak \surname Mody\orcid{0000-0001-9697-2258} \email p.p.mody@lumc.nl \\  
	\addr Department of Radiology, Leiden University Medical Center, Leiden, The Netherlands \\
        \addr HollandPTC consortium – Erasmus Medical Center, Rotterdam, Holland Proton Therapy Centre, Delft, Leiden University Medical Center, Leiden and TU Delft, Delft, The Netherlands
    \AND
    \firstname Nicolas \surname F. Chaves-de-Plaza\orcid{0000-0003-4971-3151} \email n.f.chavesdeplaza@tudelft.nl \\
    \addr Computer Graphics and Visualization Group , EEMCS, TU Delft, Delft, The Netherlands
    \AND
    \firstname Chinmay \surname Rao\orcid{0000-0002-2472-2409} \email c.s.rao@lumc.nl \\
    \addr Department of Radiology, Leiden University Medical Center, Leiden, The Netherlands
    \AND
    \firstname Eleftheria \surname Astrenidou \email e.astreinidou@lumc.nl \\
    \addr Department of Radiation Oncology, Leiden University Medical Center, Leiden, The Netherlands
    \AND
    \firstname Mischa \surname de Ridder\orcid{0000-0002-2530-3038} \email m.deridder-5@umcutrecht.nl \\
    \addr Department of Radiation Oncology, University Medical Center, Utrecht, The Netherlands
    \AND
    \firstname Nienke \surname Hoekstra\orcid{0000-0001-7355-6219} \email n.hoekstra@lumc.nl \\
    \addr Department of Radiation Oncology, Leiden University Medical Center, Leiden, The Netherlands
    \AND
    \firstname Klaus \surname Hildebrandt\orcid{0000-0002-9196-3923} \email k.a.hildebrandt@tudelft.nl \\
	\addr Computer Graphics and Visualization Group , EEMCS, TU Delft, Delft, The Netherlands
    \AND
    \firstname Marius \surname Staring\orcid{0000-0003-2885-5812} \email m.staring@lumc.nl \\
    \addr Department of Radiology, Leiden University Medical Center, Leiden, The Netherlands \\
    \addr Department of Radiation Oncology, Leiden University Medical Center, Leiden, The Netherlands   
}

\begin{document}

% top matter
\maketitle

% abstract
\begin{abstract}%   <- trailing '%' for backward compatibility of .sty file
	Increased usage of automated tools like deep learning in medical image segmentation has alleviated the bottleneck of manual contouring. This has shifted manual labour to quality assessment (QA) of automated contours which involves detecting errors and correcting them. A potential solution to semi-automated QA is to use deep Bayesian uncertainty to recommend potentially erroneous regions, thus reducing time spent on error detection. Previous work has investigated the correspondence between uncertainty and error, however, no work has been done on improving the ``utility" of Bayesian uncertainty maps such that it is only present in inaccurate regions and not in the accurate ones. Our work trains the FlipOut model with the Accuracy-vs-Uncertainty (AvU) loss which promotes uncertainty to be present only in inaccurate regions. We apply this method on datasets of two radiotherapy body sites, c.f. head-and-neck CT and prostate MR scans. Uncertainty heatmaps (i.e. predictive entropy) are evaluated against voxel inaccuracies using Receiver Operating Characteristic (ROC) and Precision-Recall (PR) curves. Numerical results show that when compared to the Bayesian baseline the proposed method successfully suppresses uncertainty for accurate voxels, with similar presence of uncertainty for inaccurate voxels. Code to reproduce experiments is available at  \url{https://github.com/prerakmody/bayesuncertainty-error-correspondence}. 
\end{abstract}

% keywords

%%%%%%%%%%%%%%%%%%%%%%%%%%%%%%%%%%%%%%%%%%%%%%%%%%%%%%%%%%%%%%%%%%%%%%%%%%%
% Introduction
%%%%%%%%%%%%%%%%%%%%%%%%%%%%%%%%%%%%%%%%%%%%%%%%%%%%%%%%%%%%%%%%%%%%%%%%%%%
% Introduction (or first section)
% Intro-1-Introduction of QA requirements
% Intro-2-Error Detection vs Error Correction
% Intro-3-Bayesian DL
% Intro-4-AvU loss
% Intro-5-Previous work

\section{Introduction}
% \setkeys{Gin}{draft}

\label{sec:introduction}
% Intro-1-Introduction of QA requirements
In recent years, deep learning models are being widely used in radiotherapy for the task of medical image segmentation. Although these models have been shown to accelerate clinical workflows \citep{clinical_time_prostate,clinical_time_hn}, they still commit contouring errors \citep{clinical_errors_hn}. Thus, a thorough quality assessment (QA) needs to be conducted, which places a higher time and manpower requirement on clinical resources. This creates a barrier to the adoption of such deep learning models \citep{clinical_accuracy_barrier}. Moreover, it also creates an obstacle for adaptive radiotherapy (ART) workflows, which have been shown to improve a patient's post-radiation quality-of-life \citep{clinical_mri_adaptive}. This \editeddd{obstacle arises} due to ART's need of regular contour updates. Currently, commercial auto-contouring tools do not have the ability to assist with quick identification and rectification of potentially erroneous predictions \citep{clinical_errors_hn,clinical_accuracy_barrier}. 

% Intro-2-Error Detection vs Error Correction
Quality assessment (QA) of incorrect contours would require two steps -- 1) error detection and 2) error correction \citep{chavesdeplaza2022}. Currently, errors are searched for by manual inspection and then rectified using existing contour editing tools. Error detection could be semi-automated by recommending either potentially erroneous slices of a 3D scan \citep{unc_sliceproposal}, or by highlighting portions of the predicted contours \citep{unc_pixelproposal} or blobs \citep{unc_blobproposal}. Upon detection of the erroneous region, the contours could be rectified using point or scribble-based techniques \citep{interactive_deepigeos,interactive_justintime} in a manner that adjacent slices are also updated. For this work, we will focus on error detection.

% Intro-3-Bayesian DL
Various approaches to error detection have suggested using Bayesian Deep Learning (BDL) and the uncertainty that it can produce as a method to capture potential errors in the predicted segmentation masks \citep{unc_sliceproposal,unc_pixelproposal,unc_blobproposal,unc_visualdetection_retinopathy,unc_visual_rtmri,unc_visual_cardiac,unc_prccurves}. Although such works established the potential usage of uncertainty in the QA of predictions, it may not be sufficient in a clinical workflow that relies on pixel-wise uncertainty as a proxy for error detection. In our experiments with deep Bayesian models, we observed that the relationship between prediction errors and uncertainty is sub-optimal, and hence has low clinical ``utility". Ideally, for semi-automated contour QA, the uncertainty should be present only in inaccurate regions and not in the accurate ones. At times, literature usually refers to this as uncertainty calibration \citep{kumar2019verified, loss_avu, zhang2020UncCalib2, unc_prccurves, gruber2022UncCalib1}, but we find this term incorrect as historically, calibration is referred to in context of probabilities of a particular event \citep{history_calib_1982}. Thus, we believe it is semantically incorrect to say uncertainty calibration and instead propose to use the term uncertainty-error correspondence.
% These methods use uncertainty as a reference to find the most erroneous slices \cite{unc_sliceproposal} and blobs \cite{unc_blobproposal,unc_prccurves}, as an input to an ancillary neural network \cite{unc_pixelproposal} to detect erroneous voxels, or simply for visual quality assessment (QA) on the predictions \cite{unc_visualdetection_retinopathy,unc_visual_rtmri,unc_visual_cardiac}

% Intro-4-AvU loss (fig)
\begin{figure}[tb]
    \includegraphics[width=\textwidth]{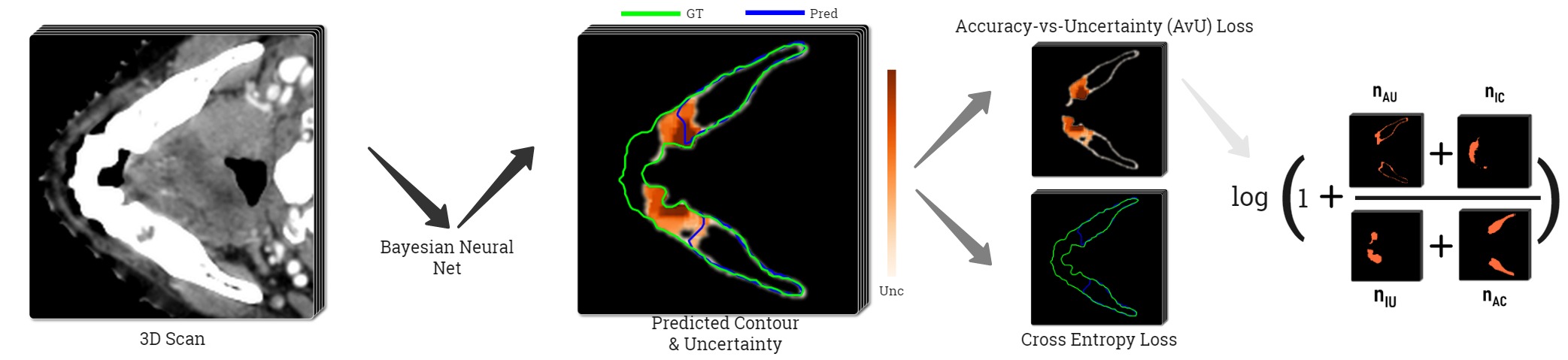}
\caption{Method overview - A 3D medical scan (e.g. CT/MR) is input into a UNet-based Bayesian neural net to produce both predicted contours (\textit{Pred}) and predictive uncertainty (\textit{Unc}). While the cross-entropy loss is used to improve segmentation performance, the Accuracy-vs-Uncertainty (AvU) loss is used to improve uncertainty-error correspondence. The AvU loss is computed by comparing the prediction with the ground truth (\textit{GT}) at a specific uncertainty threshold using four terms: count of accurate-and-certain ($n_\mathrm{AC}$), accurate-and-uncertain ($n_\mathrm{AU}$), inaccurate-and-certain ($n_\mathrm{IC}$) and inaccurate-and-uncertain ($n_\mathrm{IU}$) voxels.}
\label{fig:abstract_visual}
\end{figure}

% Intro-4-AvU loss
To create a Bayesian model that is incentivized to produce uncertainty only in inaccurate regions, we use the Accuracy-vs-Uncertainty (AvU) metric \citep{eval_avu} and its probabilistic loss version \citep{loss_avu} during training of a UNet-based  Bayesian model \citep{model_flipout}. This loss promotes the presence of both \textbf{a}ccurate-if-\textbf{c}ertain ($\mathrm{n}_{ac}$) as well as \textbf{i}naccurate-if-\textbf{u}ncertain ($\mathrm{n}_{iu}$) voxels in the final prediction (\blueautoref{fig:abstract_visual}). With uncertainty present only around potentially inaccurate regions, one can achieve improved synergy between clinical experts and their deep learning tools during the QA stage. Our work is the first to use the AvU loss in a dense prediction task like medical image segmentation and also with datasets containing natural and not synthetic variations as was previously done \citep{loss_avu}. This work extends our conference paper \citep{mody_miccai} with additional datasets, experiments and metrics. There, we adapt the original AvU loss by considering the full theoretical range of uncertainties in the loss, rather than one extracted from the validation dataset \citep{loss_avu}. For our work we use the predictive entropy as an uncertainty metric \citep{thesis_unc}.

% Intro-5-Previous work
Several other approaches have been considered in context of uncertainty, for e.g. ensembles, test time augmentation (TTA) and model calibration. While ensembles of models have good segmentation performance \citep{mehrtash2020confidence, unc_visual_cardiac}, they are parameter heavy. TTA \citep{wang2019aleatoricBrainSeg, hekler2023TtaFundus} performs inference by modulating a models inputs, but does not perform additional training, so may be unable to transcend its limitations. Calibration techniques attempt to make predictions less overconfident \citep{guo2017calibration, pereyra2017ECP, muller2019vanillaLS, mukhoti2020calibrating, islam2021spatially, murugesan2023calibrating}, however they do not explicitly align model errors with uncertainty. All the above methods are benchmarked on the truthfulness of their output probabilities (when compared against voxel accuracies) using metrics like expected calibration error (ECE). However, a model with lower ECE than its counterparts may not necessarily have higher uncertainty-error correspondence.
% Focal, LS, SVLS, MbLS, ECP, TTA
% 2017:ECP, 2019:LS, 2020:Focal, 2021: SVLS, 2023: MbLS 
% ECE: model calibration error is inversely proportional to the alignment of a models output probabilities with its pixel-wise accuracy

% Intro-6-Evaluation
Finally, to evaluate calibrative and uncertainty-error correspondence metrics, one needs to compute the ``true" inaccuracy map. Similar to our conference paper \citep{mody_miccai} and inspired by \cite{unc_pixelproposal}, we classify inaccuracies of predicted voxel maps into two categories: ``errors" and ``failures" (see \refappendix{sec:appendix_a}). Segmentation ``errors" are those inaccuracies which are considered an artifact similar to inter-observer variation, a phenomenon common in medical image segmentation \citep{clinical_iov_oar, clinical_iov_target}. Thus, we consider these smaller inaccuracies to be accurate in our computations, under the assumption they do not require clinical intervention. In the context of contour QA, such voxels should ideally be certain. Hence, only the segmentation ``failures" are a part of the ``true" inaccuracy map used to calculate the calibrative and uncertainty-error correspondence metrics.
% \newline 

To summarize, our contributions are as follows:
\begin{itemize} 
    \item For the purpose of semi-automated quality assessment of predicted contours, we aim to improve uncertainty-error correspondence (unc-err) in a Bayesian medical image segmentation setting, pioneering this in the context of radiation therapy. Specifically, we propose using the loss form of the Accuracy-vs-Uncertainty (AvU) metric while training a deep Bayesian segmentation model.
    \item We compare our Bayesian model with the AvU loss against an ensemble of deterministic models, five approaches employing calibration-based losses and also test time augmentation. We also perform an architectural comparison by comparing models with Bayesian convolutions placed in either the middle layers or decoder layers of a deep segmentation model.
    \item We benchmark unc-err of the segmentation models on both in- and out-of-distribution radiotherapy datasets containing head-and-neck CT and Prostate MR scans. Models are benchmarked on these datasets across discriminative, calibrative and uncertainty-error correspondence metrics.
\end{itemize}

%%%%%%%%%%%%%%%%%%%%%%%%%%%%%%%%%%%%%%%%%%%%%%%%%%%%%%%%%%%%%%%%%%%%%%%%%%%
% Related works
%%%%%%%%%%%%%%%%%%%%%%%%%%%%%%%%%%%%%%%%%%%%%%%%%%%%%%%%%%%%%%%%%%%%%%%%%%%
% Make sure to put your work into context and include apporpriate citations.
% We do not have limits on citation counts.
% 1 - Epistemic/Aleatoric
% 2 - Uncertainty Metric (cancelled in MELBA2)
% 3 - Refinement
% 4 - Calibration

\section{Related Works}
% 1 - Epistemic/Aleatoric
\subsection{Epistemic and aleatoric uncertainty}
Recent years have seen an increase in work that utilizes probabilistic modeling in deep medical image segmentation. The goal has been to account for uncertainty due to noise in the dataset (\textit{aleatoric uncertainty}) as well as in the limitations of the predictive models learning capabilities (\textit{epistemic uncertainty}). Noise in medical image segmentation refers to factors like inter- and intra- annotator contour variation \citep{clinical_iov_oar, clinical_iov_target} due to factors such as poor contrast in medical scans. Works investigating aleatoric uncertainty model the contour diversity in a dataset by either placing Gaussian noise assumptions on their output \citep{unc_aleatoric_sns} or by assuming a latent space in the hidden layers and training on datasets containing multiple annotations per scan \citep{unc_alea_learn}. A popular and easy-to-implement approach to model for aleatoric uncertainty is called test-time augmentation (TTA) \citep{unc_tta}. Here, different transformations of the image are passed through a model, and the resulting outputs are combined to produce both an output and its associated uncertainty. 

In contrast to aleatoric uncertainty, epistemic uncertainty could be used to identify scans (or parts of the scan) that are very different from the training dataset. Here, the model is unable to make a proper interpolation from its existing knowledge. Methods such as ensembling \citep{mehrtash2020confidence} and Bayesian posterior inference (e.g., Monte-Carlo DropOut, Stochastic Variational Inference) \citep{unc_pixelproposal,unc_blobproposal,unc_sliceproposal,unc_visualdetection_retinopathy,unc_visual_rtmri,unc_triaging, loss_avu} are common methods to model epistemic uncertainty in neural nets. While Bayesian modeling is a more mathematically motivated and hence, principled approach to estimating uncertainty, ensembles have been motivated by the empirically-proven concept of bootstrapping. In contrast to Bayesian models where the perturbation is modelled by placing distributions on weights, ensembles use either different model weight initializations, or different subsets of the training data. In Bayesian inference techniques, perturbations are introduced in the models activation or weight space. Dropout \citep{model_dropout} and DropConnect \citep{model_dropconnect} are popular techniques that apply the Bernoulli distribution on these spaces. Stochastic variational inference (SVI) is another type of weight space perturbation that usually assumes the more expressive Gaussian distribution on the weights. Bayes by Backprop \citep{model_bbb} and its resource-efficient variant such as FlipOut \citep{model_flipout} are examples of SVI. For our work, we consider approaches that are designed for both epistemic uncertainty (Ensembles and SVI models) as well as aleatoric uncertainty (TTA).

% 3 - Refinement
\subsection{Uncertainty use during training}
Other works also use the uncertainty from a base segmentation network to automatically refine its output using a follow-up network. This refinement network can be graphical \citep{unc_refine_graph} or simply convolutional \citep{unc_pixelproposal}. Uncertainty can also be used in an active learning scenario, either with \citep{unc_refine_deepedit} or without \citep{unc_refine_curriculum} interactive refinement. Shape-based features of uncertainty maps have also been shown to identify false positive predictions \citep{unc_refine_shapefeat}. Similarly, we too use uncertainty in our training regime, but with the goal of promoting uncertainty only in those regions which are inaccurate, an objective not previously explored in medical image segmentation.

% 4 - Calibration
\subsection{Model calibration}
In context of segmentation, model calibration error is inversely proportional to the alignment of a models output probabilities with its pixel-wise accuracy. Currently there is no proof that reduction in model calibration error leads to improved uncertainty-error correspondence. However, a weak link can be assumed since both are derived from a models output probabilities. It is well known that the probabilities of deterministic models trained on the cross entropy (CE) loss are not well calibrated \citep{guo2017calibration}. This means that they are overconfident on incorrect predictions and hence fail silently\sout{. This}\editeddd{, which} is an undesirable trait in context of segmentation QA and needs to be resolved.

To abate this \editeddd{overconfidence} issue, methods such as post-training model calibration (or temperature scaling) \citep{guo2017calibration, ding2021local, calibration_improve}, ensembles \citep{mehrtash2020confidence, unc_shift_ovadia2019}, calibration-focused training losses \citep{pereyra2017ECP, mukhoti2020calibrating, murugesan2023calibrating, murugesan2023trust} and calibration-focused training targets \citep{muller2019vanillaLS, islam2021spatially} have been shown to improve model calibration for deterministic models. Temperature scaling, a post-training model calibration technique, has been shown to perform poorly in out-of-domain (OOD) settings \citep{unc_shift_ovadia2019}, relies wholly on an additional validation dataset and/or needs explicit shape priors \citep{calibration_improve}.\sout{ Finer}\editeddd{Local} temperature scaling techniques have been proposed that calibrate on the image or pixel level \citep{ding2021local}, however they are still conceptually similar to the base method and are hence plagued by the same concerns. Others \citep{calibration_improve} used a shape prior module for out-of-domain robustness, but they only introduced synthetic textural variations in their work. 

Another approach to model calibration is to regularize a model during train to promote uncertainty. For e.g. the ECP \citep{pereyra2017ECP} technique explicitly adds the negative entropy to the training loss. Conversely, the Focal loss \citep{lin2017focal,mukhoti2020calibrating}\sout{ achieves this}\editeddd{attempts to calibrate a model} implicitly by assigning lower weights \editeddd{(during training)} to more confident predictions. Other methods smooth the hard targets of the ground truth towards a uniform distribution in the limit. For e.g. Label Smoothing \citep{szegedy2016rethinking, muller2019vanillaLS}\sout{ does this by modifying}\editeddd{ modifies} the class distribution of a pixel \editeddd{by calculating a weighted average (using parameter $\alpha$) between the hard target and a uniform distribution}. \editeddd{On the other hand,} Spatial Varying Label Smoothing (SVLS) \citep{islam2021spatially} modifies a \editeddd{pixel's} class allocation by considering classes around it. \sout{To avoid excessively making the models predictions uniform,} Margin-based Label Smoothing (MBLS) \citep{liu2022devil, murugesan2023calibrating} \editeddd{reformulates the above approaches by showing that they essentially perform loss optimization where an equality constraint is applied on a pixels logits. MBLS} attempts to achieve the best discriminative-calibrative trade-off \editeddd{by softening this equality constraint. They subtract the max logit of a pixel with its other logits and} only penalize those logit distances that are greater than a predetermined margin. Others extend \sout{this}\editeddd{the MBLS} framework by \sout{further tuning}\editeddd{either learning} class-specific weights \editeddd{for the equality constraint} \citep{liu2023class} or reformulating SVLS to a \sout{similar }formulation \editeddd{similar to MBLS} \citep{murugesan2023trust}. Although these methods attempt to make models less overconfident, they do not explicitly align a model's error to its uncertainty.

There also exist other approaches to model calibration for e.g.,  multi-task learning \citep{karimi2022improving}, mixup augmentation \citep{thulasidasan2019mixup} and shape priors \citep{karimi2019accurate}. Multi-task learning requires additional data that may not always be present, while mixup creates synthetic samples which are not representative of the real data distribution. Finally, shape priors may not be applicable to tumors with variable morphology.

Model calibration techniques are evaluated by metrics like Expected Calibration Error (ECE) and its variants \citep{calibration_thace}, however others have also proposed terms like Uncertainty-Calibration Error (UCE) \citep{laves2019well, PPR:PPR517139}. While ECE evaluates the equivalency between accuracy and predicted probability, UCE compares inaccuracy and uncertainty. However, while it is semantically appropriate to expect an average probability of $p\ (0 \leq p \leq 1$) to give the same average accuracy (i.e., the mathematical formulation of ECE), the same is not appropriate for inaccuracy and uncertainty $u\ (0 \leq u \leq 1$). Hence, UCE is not applicable to our work.

To conclude, the issue with each of the aforementioned techniques for epistemic, aleatoric and calibrative modeling is that they do not explicitly train the model to develop an innate sense of potential errors on a given segmentation task. Given that this is the primary requirement from a contour QA perspective, these models may be unable to have good uncertainty-error correspondence.

% A methodological, model, or similar section often comes here.
\section{Methods}
\subsection{Neural Architecture}
We adopt the OrganNet2.5D neural net architecture \citep{model_organnet2.5d} which is a standard encoder-decoder model connected by four middle layers. It contains both 2D and 3D convolutions in the encoder and decoder as well as hybrid dilated convolutions (HDC) in the middle. This network performs fewer pooling steps to avoid losing image resolution and instead uses HDC to expand the receptive field. To obtain uncertainty corresponding to the output, we add stochasticity to the deterministic convolutional operations by replacing them with Bayesian convolutions \citep{model_bbb,model_flipout}. We experiment with replacing deterministic layers in both the HDC as well as the decoder layers to understand the effect of placement.

In a Bayesian model, a prior distribution is placed upon the weights and is then updated to a posterior distribution on the basis of the training data. During inference (\blueautoref{eq:1_variation_inference}), we sample from this posterior distribution $p(W|D)$ to estimate the output distribution $p(y|x,D)$ with $x$, $y$ and $W$ being the input, output and neural weight respectively:
\begin{equation}
    p(y|x,D) = \mathbbm{E}_{W \sim{p(W|D)}} \Bigr[p(y|x,W)\Bigr].
    \label{eq:1_variation_inference}
\end{equation}
This work uses a Bayesian posterior estimation technique called stochastic variational inference, where instead of finding the true, albeit intractable posterior, it finds a distribution close to it. We chose FlipOut-based \citep{model_flipout} convolutions which assume the distribution over the neural weights to be a Gaussian and are factorizable over each hidden layer. Pure variational approaches would need to sample from this distribution for each element of the mini-batch \citep{model_bbb}. However, the FlipOut technique only samples once and multiplies that random sample with a Rademacher matrix, making the forward pass less computationally expensive.

\subsection{Training Objectives}

In this section , we use a notation format, where capital letters denote arrays while non-capital letters denote scalar values. 

\subsubsection{Segmentation Objective}
Upon being provided a 3D scan as input, our segmentation model predicts for each class $c \in C$, a 3D probability map $\hat{P_c}$ of the same size. Each voxel $i \in N$ has a predicted probability vector $\hat{P^i}$ containing values \editeddd{$\hat{p_c^i}$} for each class that sum to 1 (due to softmax). To calculate the predicted class of each voxel $\hat{y^i}$, we do:
\begin{align}
        \hat{y^i} &= \operatorname*{argmax}_{c \in C}\hat{p_c^i}.
\end{align} \label{eq:2_argmax}

To generate a training signal, the predicted probability vector $\hat{P^i}$ is compared to the corresponding one-hot vector $Y^i$ in the gold standard 3D annotation mask. \editeddd{$Y^i$ is composed of $y^i_c \in \{0,1\}$}. 
Inspired by \citep{taghanaki2019combo,yeung2022unified}, we re-frame the binary cross-entropy loss (\blueautoref{eq:3_cross_entropy}), as penalizing both the foreground \editeddd{($y_{c}^{i}=1$)} and background \editeddd{(($1-y_c^i)=1$)} voxels of the probability maps of each class with a weight $w_c$:

% \begin{equation}
%     \begin{aligned}
%     L_\mathrm{CE} &= - \frac{1}{|C|}\left(\displaystyle\sum_{c \in C} w_c \Bigr[      
%     \displaystyle\sum_{i \in N}\bigm(
        
%         y_{c}^{i} \ln(\hat{p_c^i})
%         + 
%         (1-y_c^i) \ln(1 - \hat{p_c^i})
        
%     \bigm)\Bigr]\right).
%     \end{aligned}
%     \label{eq:3_cross_entropy}
% \end{equation}
\begin{equation}
    \begin{aligned}
    L_\mathrm{CE} &= - \frac{1}{|C|} \left( \sum_{c \in C} w_c \left[      
    \sum_{i \in N} \left(
        y_{c}^{i} \ln(\hat{p}_c^i)
        + 
        (1-y_c^i) \ln(1 - \hat{p}_c^i)
    \right) \right] \right).
    \end{aligned}
    \label{eq:3_cross_entropy}
\end{equation}

Note, we do not utilize the DICE loss for training as it has been shown to have lower model calibration metrics \citep{mody2022comparing}. Also, since the CE loss is susceptible to fail during a class-imbalance, we use its weighted version.

\subsubsection{Uncertainty Objective}
In a Bayesian model, multiple forward passes ($m \in M$) are performed and the output 3D probability maps $\hat{(P_c)_m}$ of each pass are averaged to output $\hat{P_c}$ (\blueautoref{eq:1_variation_inference}). Using $\hat{P_c}$, we can calculate a host of statistical measures like entropy, mutual information and variance. We chose entropy as it has been shown to capture both epistemic uncertainty, which we explicitly model in FlipOut layers, as well as aleatoric uncertainty, which is implicitly modeled due to training data \citep{thesis_unc}. We use the predicted class probability vector $\hat{P^i}$ for each voxel and calculate its (normalized) entropy $u^i$:

\begin{equation}
    \begin{aligned}
      u^i &= -\frac{1}{\ln (|C|)}\sum_{c \in C} \hat{p_c^i} \ln (\hat{p_c^i}).
      \label{eq:4_entropy}
    \end{aligned}   
\end{equation}

Since we have access to the gold standard annotation mask, each voxel has two properties: accuracy and uncertainty. Accuracy is determined by comparing the gold standard class $y^i$ to the predicted class $\hat{y^i}$. We use this to classify them in four different categories represented by $n_\mathrm{ac}$, $n_\mathrm{au}$, $n_\mathrm{ic}$ and $n_\mathrm{iu}$, where $n$ stands for the total voxel count and $a$, $i$, $u$, $c$ represent the \textbf{a}ccurate, \textbf{i}naccurate, \textbf{u}ncertain and \textbf{c}ertain voxels. A visual representation of these terms can be seen in \blueautoref{fig:abstract_visual}. Here, a voxel is determined to be certain or uncertain on the basis of a chosen uncertainty threshold $t \in T$ where the maximum value in $T$ is the maximum theoretical uncertainty threshold \citep{mody_miccai}. The aforementioned four terms are the building blocks of the Accuracy-vs-Uncertainty (AvU) metric \citep{eval_avu} as shown in \blueautoref{eq:5_avu} - \blueautoref{eq:6_avu_terms} and it has a range between [0,1]. A higher value indicates that uncertainty is present less in accurate regions and more in inaccurate regions, thus improving the ``utility" of uncertainty as a proxy for error detection.

\begin{align}
    \mathrm{AvU^t} &= \frac{n_{\mathrm{ac^t}} + n_{\mathrm{iu^t}}}{n_{\mathrm{ac^t}} + n_{\mathrm{au^t}} + n_{\mathrm{ic^t}} + n_{\mathrm{iu^t}}}\label{eq:5_avu} \\
    n_\mathrm{ac}^t &= \displaystyle\sum_{i \in \Bigr\{ \substack{y_i = \hat{y_i} \ \& \\ u_i \leq t} \Bigr\}} 1, \qquad
    n_\mathrm{au}^t = \displaystyle\sum_{i \in \Bigr\{ \substack{y_i = \hat{y_i} \ \& \\ u_i > t} \Bigr\}} 1 \\
    n_\mathrm{ic}^t &= \displaystyle\sum_{i \in \Bigr\{ \substack{y_i \neq \hat{y_i} \ \& \\ u_i \leq t} \Bigr\}} 1, \qquad 
    n_\mathrm{iu}^t = \displaystyle\sum_{i \in \Bigr\{ \substack{y_i \neq \hat{y_i} \ \& \\ u_i > t} \Bigr\}} 1\label{eq:6_avu_terms}
\end{align}

To maximize AvU for a neural net, one can turn it into a loss metric to be minimized. As done in \cite{loss_avu} for an image classification setting, we minimize its negative logarithm (\blueautoref{eq:7_avu_diff}) to improve mathematical stability of gradient descent. However, the AvU metric, as defined above, is not differentiable \editeddd{with respect to the neural net's weights. This is due to all its constituent terms being produced either due to thresholding or max operations which introduce discontinuities that disrupt gradient flows.}. \sout{This is because the model's outputs are simply used to create a mask and hence no backpropagation can take place.} The AvU metric is made differentiable by instead using \editeddd{the uncertainty $u^i$ derived from $\hat{P^i}$ (\blueautoref{eq:4_entropy}), thus allowing for gradient flows}. \sout{by instead using the maximum value from each voxel's predicted probability vector $p^i = \max(P^i)$}. Also, a smooth non-linear operation i.e., \textit{tanh} is used to constrain its value (\blueautoref{eq:8_avu_diff_terms_accurate}). \editeddd{The differentiable uncertainty term is multiplied by other scalar weighing terms c.f. the maximum probability ($\hat{p^i} = \max(\hat{P^i})$) and accuracy/inaccuracy mask for a voxel. All these operations together allow us to calculate proxy values for $n_\mathrm{ac}$, $n_\mathrm{au}$, $n_\mathrm{ic}$ and $n_\mathrm{iu}$.} In addition, rather than evaluating the loss at a single uncertainty threshold, we integrate over the theoretical range of the uncertainty metric. \editeddd{Thresholding is done by once again multiplying the uncertainty value with a binary mask.} The benefits of \sout{this} thresholding were shown in our conference paper \citep{mody_miccai}:

\begin{align}
\begin{split}
        L_{\mathrm{AvU^t}} &= - \ln\left(1 + \frac{n_{\mathrm{au}}^t + n_{\mathrm{ic}}^t}{n_{\mathrm{ac}}^t + n_{\mathrm{iu}}^t}\right), \\
        L_{\mathrm{AvU}} &= \frac{1}{T} \displaystyle\sum_{t \in T} L_{\mathrm{AvU^t}},
\end{split}
\label{eq:7_avu_diff}
\end{align}
where
\begin{equation}
        \begin{split}
            n_\mathrm{ac}^t &= \displaystyle\sum_{i \in \Bigr\{ \substack{y_i = \hat{y_i} \ \& \\ u_i \leq t} \Bigr\}} \hat{p^i} \cdot (1 - \mathrm{tanh}(u^i)), \\
            n_\mathrm{ic}^t &= \displaystyle\sum_{i \in \Bigr\{ \substack{y_i \neq \hat{y_i} \ \& \\ u_i \leq t} \Bigr\}} (1-\hat{p^i}) \cdot (1-\mathrm{tanh}(u^i)),    
        \end{split}
        \qquad
        \begin{split}
            n_\mathrm{au}^t &= \displaystyle\sum_{i \in \Bigr\{ \substack{y_i = \hat{y_i} \ \& \\ u_i > t} \Bigr\}} \hat{p^i} \cdot \mathrm{tanh}(u^i) \\
            n_\mathrm{iu}^t &= \displaystyle\sum_{i \in \Bigr\{ \substack{y_i \neq \hat{y_i} \ \& \\ u_i > t} \Bigr\}} (1-\hat{p^i}) \cdot \mathrm{tanh}(u^i).    
        \end{split}
\label{eq:8_avu_diff_terms_accurate}
\end{equation}

Finally, the total loss $L$ combines the segmentation and uncertainty loss as:
\begin{equation}
    \begin{aligned}
        L =L_{\mathrm{CE}} + \alpha \cdot L_{\mathrm{AvU}}.
    \end{aligned}
\end{equation}

\subsection{Evaluation}

\subsubsection{Discriminative and Calibration Evaluation}
We evaluate all models on the DICE coefficient for discriminative performance. Calibration is evaluated using the Expected Calibration Error (ECE) \citep{guo2017calibration}. Numerical results are compared with the Wilcoxon signed-ranked test where a p-value $\leq$ 0.05 is considered significant.

\subsubsection{Uncertainty Evaluation}
As the model is trained on the Accuracy-vs-Uncertainty (AvU) metric, we calculate the AvU scores up to the maximum normalized uncertainty of the validation dataset. A curve with the AvU score on the y-axis and the uncertainty threshold on the x-axis is made and the area-under-the-curve (AUC) for each scan is calculated. AUC scores provide us with a summary of the model performance regardless of the uncertainty threshold, and hence we use it to compare all models.   

The AvU metric outputs a single scalar value for the whole scan and does not offer much insight into the differences in uncertainty coverage between the accurate and inaccurate regions. To abate this, we compare the probability of \textbf{u}ncertainty in \textbf{i}naccurate regions  $p(u|i)$ to the probability of \textbf{u}ncertainty in \textbf{a}ccurate regions $p(u|a)$. Let us plot $p(u|i)$ and $p(u|a)$ on the y-axis and x-axis of a graph respectively, and define $n_\mathrm{iu}$, $n_\mathrm{au}$, $n_\mathrm{ac}$ and $n_\mathrm{ic}$, as the count of true positives, false positives, true negatives and false negatives respectively. Thus, $p(u|i)$ is the true positive rate and $p(u|a)$ is the false positive rate. Computing this at different uncertainty thresholds provides us with the Receiver Operating Characteristic (ROC) curve, which we call the uncertainty-ROC curve \citep{model_dropconnect}. 

Given that ROC curves are biased in situations with class imbalances between positive (inaccurate voxels) and negative (accurate voxels) classes, we also compute the precision-recall curves \citep{unc_prccurves}.Here, precision is the probability of inaccuracy given uncertainty $p(i|u)$ and recall is the probability of uncertainty given inaccuracy $p(u|i)$. Note, that the precision-recall curves do not make use of $n_\mathrm{ac}$, which can be high in count for a well-performing model.

Finally, to calculate the calibrative and uncertainty-correspondence metrics, we need an inaccuracy map. We use an inaccuracy map based on the concept of segmentation ``failures" and ``errors" (\refappendix{sec:appendix_a}). To do this, we perform a morphological opening operation using a fixed kernel size of (3,3,1).

\section{Experiments and Results}
% \setkeys{Gin}{draft}

\subsection{Datasets}
\subsubsection{Head-and-Neck CT}
Our first dataset contained Head and Neck CT scans of patients from the RTOG 0522 clinical trial \citep{dataset_rtog}. The annotated data, which had been collected from the MICCAI2015 Head and Neck Segmentation challenge, contained 33 CT scans for training, 5 for validation and 10 for testing \citep{dataset_miccai}. We further expanded the test dataset with annotations of 8 patients belonging to the RTOG trial from the DeepMindTCIA dataset (DTCIA) \citep{dataset_deepmind}. This dataset included annotations for the mandible, parotid glands, submandibular glands and brainstem. Although there were annotations present for the optic organs, we ignored them for this analysis as they are smaller compared to other organs and require special architectural design choices. Since the train and test patients came from the same study, we considered this as an in-distribution dataset. We also tested our models on the STRUCTSeg (50 scans) dataset \citep{dataset_structseg}, hereby shortened as STRSeg. While the RTOG dataset contained American patients, the STRSeg dataset was made up of Chinese patients and hence considered out-of-distribution (OOD) in context of the training data. The uncertainties of this dataset were evaluated to a value of 0.4 since that is the maximum empirical normalized entropy.

\subsubsection{Prostate MR}
Our second dataset contained MR scans of the prostate for which we use the ProstateX repository \citep{dataset_prostatex} containing 66 scans as the training dataset. The Medical Decathlon (Prostate) dataset with 34 scans \citep{dataset_meddec} and the PROMISE12 repository with 50 scans \citep{dataset_promise12} served as our test dataset. The Medical Decathlon dataset (abbreviated as PrMedDec henceforth) contained scans from the same clinic as the ProstateX training dataset. We combined the Peripheral Zone (PZ) and Transition Zone (TZ) from the MedDec dataset into 1 segmentation mask. The PROMISE12 dataset (abbreviated as PR12) was chosen for testing since literature \citep{mehrtash2020confidence} has shown lower performance on it and hence it serves as a good candidate to evaluate the utility of uncertainty. This dataset is different from ProstateX due to the usage of an endo-rectal coil in many of its scans as well as the presence of gas pockets in the rectum and dark shadows due to the usage of older MR machines. Thus, although these datasets contained scans of the prostate region, there exists a substantial difference in their visual textures. The maximum empirical normalized entropy of this 2-class dataset is 1.0 and hence the uncertainty-error correspondence metrics were calculated till this value.

\subsection{Experimental Settings}
We tested the Accuracy-vs-Uncertainty (AvU) loss on four datasets containing scans of different modalities and body sites. We trained 11 models: \textit{Det} (deterministic), \textit{Det+AvU}, \textit{Ensemble}, \textit{Focal}, \textit{LS} (Label Smoothing), \textit{SVLS} (Spatially Varying Label Smoothing), \textit{MbLS} (Margin based Label Smoothing), \textit{ECP (Explicit Confidence Penalty)}, TTA (Test-Time Augmentation), \textit{Bayes} and \textit{Bayes + AvU}. As the names suggest, \textit{Bayes} and \textit{Bayes + AvU} are Bayesian versions of the deterministic OrganNet2.5D model \citep{model_organnet2.5d}. The baseline \textit{Bayes} model contained Bayesian convolutions in its middle layers and was trained using only the cross-entropy (CE) loss. The \textit{Bayes + AvU} was trained using both the CE and Accuracy-vs-Uncertainty (AvU) loss. Two additional Bayesian models were trained which tests if the placement of the Bayesian layers had any effect: \textit{BayesH} and \textit{BayesH + AvU}. Here, \textit{BayesH} refers to the Bayesian model with Bayesian layers in the head of the model (i.e the decoder). Results for these models can be found in \refappendix{sec:app_bayesh}.

The \textit{Ensemble} was made of $M=5$ deterministic models with different initializations \citep{unc_shift_ovadia2019}. For TTA, we applied Gaussian noise and random pixel removals for $M=5$ times each and then averaged their outputs. The hyperparameters of the other models were chosen on the basis of the best discriminative, calibrative and uncertainty-error correspondence metrics on the validation datasets (\ref{sec:appendix_c}). For the calibration focused methods we used the following range of hyperparameters: Focal ($\gamma=1,2,3$), MbLS ($m=8,10,20,30$) for head-and-neck CT, MbLS ($m=3,5,8,10$) for prostate MR, LS ($\alpha=0.1, 0.05, 0.01$), SVLS ($\gamma=1,2,3$) and ECP ($\lambda=0.1, 1.0, 10.0, 100.0$) for head-and-neck CT and ECP ($\lambda=0.1, 1.0, 10.0, 100.0, 1000.0$) for prostate MR. For the AvU loss, we evaluated weighting factors in the range $[$10,100,1000,10000$]$ for the head-and-neck dataset, and $[$100,1000,10000$]$ for the Prostate dataset.

We trained our models for 1000 epochs using the Adam optimizer with a fixed learning rate of 10$^{-3}$. The deterministic model contained $\approx$ 550K parameters and thus the \textit{Ensemble} contained $\approx$ 2.75M parameters. Since the Bayesian models double the parameter count in their layers they incurred an additional parameter cost and ended up with a total of $\approx$ 900K parameters. 

% HN-Det   : 562,395
% HN-Bayes : 894,171
% HN-Head  : 584,135

% Prostate-Det      : ?  
% Prostate-Bayes    : 892,014
% Prostate-BayesHead: 578,797(sigmoid) vs 578,814 (softmax)

\subsection{Results} 
In \blueautoref{sec:results_hn} and \blueautoref{sec:results_pros} we show discriminative (DICE), calibrative (ECE) and uncertainty-error correspondence metrics (ROC-AUC, PRC-AUC) for the two datasets.

\subsubsection{Head-and-neck CT}
\label{sec:results_hn}

%%%%%%%%%%%%%%%%%%%%%%%%%%%%%%%%%%%%%%%%%%%%%%% Table-HN
% AvU = n_ac + n_iu / (N=(n_ac + n_au + n_iu + n_ic))
% ROC = p(u|i) vs p(u|a)
% PRC = p(u|i) vs p(i|u)

\begin{table}[!tb]
    \centering
    \caption{Volumetric (\textit{DICE}), calibrative (\textit{ECE}) and uncertainty-error correspondence (ROC-AUC, PRC-AUC) metrics for all models. Here, we evaluate head-and-neck (H\&N) CT test datasets which are either in-distribution (ID) or out-of-distribution (OOD). The arrows in the table header indicate whether a metric should be high ($\uparrow$) or low ($\downarrow$). Here, $^{\dag}$ and \textbf{bold} are used to indicate a statistical significance and improved results upon comparing a Bayesian model and its AvU-loss version, while \underline{underlined} numbers indicate the best value for a metric across a dataset. }
    
    \begin{tabular}{|c|l|l|l|l|l|}
    
    \hline
    \thead{Test \\ Dataset} & \thead{Model} & \thead{DICE $\uparrow$\\(x$10^{-2}$)} & \thead{ECE $\downarrow$\\(x$10^{-2}$)} & \thead{ROC-AUC $\uparrow$ \\(x$10^{-2}$)} & \thead{PRC-AUC $\uparrow$ \\(x$10^{-2}$)} \\
    \hline

    %%%%%%%%%%%% RTOG %%%%%%%%%%%%
    \hline
    \multirow{11}{*}{\parbox{2.0cm}{\centering ID \\ ------------ \\ H\&N CT \\ (RTOG)}} 
     & Det        & 84.2 ± 2.7  & 9.0 ± 2.1  & 73.0 ± 5.7 & 21.0 ± 4.8  \\
     & \editedd{Det + AvU}  & \editedd{83.8 ± 2.9} & \editedd{8.6 ± 2.7} & \editedd{73.1 ± 6.0} & \editedd{20.8 ± 4.0}  \\
     \cdashline{2-6}
     & Focal          & 84.3 ± 2.4  & 9.3 ± 1.5  & 70.3 ± 5.5 & 18.2 ± 3.2  \\
     & \editedd{ECP}  & \editedd{84.4 ± 2.3}  & \editedd{9.0 ± 2.0}  & \editedd{73.8 ± 5.4} & \editedd{21.0 ± 3.7}  \\
     & \editedd{LS}   & \editedd{83.0 ± 3.0}  & \editedd{7.5 ± 2.2}  & \editedd{62.6 ± 3.3} & \editedd{17.5 ± 4.0}  \\ 
     & \editedd{SVLS} & \editedd{84.2 ± 2.6}  & \editedd{9.0 ± 2.0}  & \editedd{70.8 ± 7.1} & \editedd{18.1 ± 3.5}  \\
     & MbLS           & 84.0 ± 2.6  & 9.2 ± 2.1  & 67.5 ± 5.7 & 19.5 ± 3.5  \\
     \cdashline{2-6}
     & \editedd{TTA}  & \editedd{84.1 ± 2.8} & \editedd{9.1 ± 2.1}   & \editedd{72.9 ± 5.9} & \editedd{20.8 ± 3.9}  \\
     \cdashline{2-6}
     & Ensemble       & \underline{85.0 ± 2.6}  & 7.8 ± 1.8  & \underline{78.6 ± 4.7} & \underline{25.7 ± 6.8}  \\
     \cdashline{2-6}
     & Bayes      & 83.9 ± 2.6 & 8.6 ± 2.1 & 74.1 ± 5.4 & 22.1 ± 3.5  \\
     & Bayes+AvU  & 83.6 ± 2.5 & \textbf{\underline{7.6 ± 2.5}}$^{\dag}$ & \textbf{76.1 ± 5.6}$^{\dag}$ & \textbf{25.1 ± 5.3}$^{\dag}$  \\

    %%%%%%%%%%%% STRSeg %%%%%%%%%%%%
    \hline
    \multirow{11}{*}{\parbox{2.0cm}{\centering OOD \\ ------------ \\ H\&N CT \\ (STRSeg)}} 
     & Det                  & 78.1 ± 4.6  & 12.9 ± 2.6 & 62.2 ± 4.5 & 24.1 ± 3.7  \\
     & \editedd{Det + AvU}  & \editedd{78.6 ± 4.7}  & \editedd{12.7 ± 3.0} & \editedd{60.8 ± 4.7} & \editedd{22.4 ± 4.1}  \\
     \cdashline{2-6}
     & Focal               & 77.2 ± 6.7  & 12.5 ± 2.9 & 57.0 ± 4.6 & 20.9 ± 4.2  \\
     & \editedd{ECP}       & \editedd{78.8 ± 4.3}  & \editedd{12.5 ± 2.6} & \editedd{61.5 ± 4.8} & \editedd{23.2 ± 3.6}  \\
     & \editedd{LS}         & \editedd{77.7 ± 6.0}  & \editedd{10.3 ± 2.9} & \editedd{56.7 ± 3.3} & \editedd{20.6 ± 4.3}  \\
     & \editedd{SVLS}      & \editedd{79.0 ± 6.0}  & \editedd{11.3 ± 2.5} & \editedd{59.9 ± 5.4} & \editedd{21.6 ± 2.7}  \\
     & MbLS       & 77.5 ± 6.3  & 13.4 ± 3.0 & 56.9 ± 5.0 & 21.5 ± 3.6  \\ 
     \cdashline{2-6}
     & \editedd{TTA}        & \editedd{78.1 ± 4.6}  & \editedd{12.7 ± 2.6} & \editedd{62.7 ± 4.6} & \editedd{24.9 ± 4.1}  \\
     \cdashline{2-6}
     & Ensemble   & \underline{78.6 ± 5.2}  & \underline{10.6 ± 2.4} & 64.7 ± 4.9 & 28.2 ± 5.1  \\
     \cdashline{2-6}
     & Bayes      & 75.0 ± 9.9  & 12.4 ± 4.0 & 64.8 ± 5.0 & 27.7 ± 5.8  \\
     & Bayes+AvU  & 76.3 ± 7.7  & \textbf{12.1 ± 3.7} & \underline{\textbf{65.8 ± 5.0}}$^{\dag}$ & \underline{\textbf{30.1 ± 6.5}}$^{\dag}$  \\
    
    \hline
    \end{tabular}
    
    \label{tab:results_table_hn}
\end{table}

%%%%%%%%%%%%%%%%%%%%%%%%%%%%%%%%%%%%%%%%%%%%%%% Table-Pros
% AvU = n_ac + n_iu / (N=(n_ac + n_au + n_iu + n_ic))
% ROC = p(u|i) vs p(u|a)
% PRC = p(u|i) vs p(i|u)
\begin{table}[tb!]
    \centering
    \caption{Volumetric (\textit{DICE}), calibrative (\textit{ECE}) and uncertainty-error correspondence (ROC-AUC, PRC-AUC) metrics for all models. Here, we evaluate Prostate MR test datasets which are either in-distribution (ID) or out-of-distribution (OOD). The arrows in the table header indicate whether a metric should be high ($\uparrow$) or low ($\downarrow$). Here, $^{\dag}$ and \textbf{bold} are used to indicate a statistical significance and improved results upon comparing a Bayesian model and its AvU-loss version, while \underline{underlined} numbers indicate the best value for a metric across a dataset.}
    
    \begin{tabular}{|c|l|l|l|l|l|}
    
    \hline
    \thead{Test \\ Dataset} & \thead{Model} & \thead{DICE $\uparrow$\\(x$10^{-2}$)} & \thead{ECE $\downarrow$\\(x$10^{-2}$)} & \thead{ROC-AUC $\uparrow$ \\(x$10^{-2}$)} & \thead{PRC-AUC $\uparrow$ \\(x$10^{-2}$)} \\
    \hline

    %%%%%%%%%%%% RTOG %%%%%%%%%%%%
    \hline
    \multirow{11}{*}{\parbox{2.0cm}{\centering ID \\ ------------ \\ Prostate MR \\ (PrMedDec)}} 
     & Det        &  84.1 ± 5.6  & 12.9 ± 6.0 &  92.5 ± 5.7 & 28.0 ± 3.7  \\
     & \editedd{Det + AvU}  &  \editedd{83.7 ± 6.8}  & \editedd{16.9 ± 8.1} &  \editedd{92.1 ± 6.8} & \editedd{28.2 ± 3.4}  \\
     \cdashline{2-6}
     & Focal                &  81.1 ± 15.4 & 10.2 ± 5.0 &  93.2 ± 5.5 & 29.3 ± 3.4  \\
     & \editedd{ECP}        &  \editedd{84.0 ± 5.5}  & \editedd{16.7 ± 7.1} &  \editedd{92.1 ± 6.0} & \editedd{27.6 ± 4.3}  \\
     & \editedd{LS}         &  \editedd{83.4 ± 7.2}  & \editedd{15.1 ± 8.6} &  \editedd{83.2 ± 7.8} & \editedd{25.1 ± 3.1}  \\ 
     & \editedd{SVLS}       &  \editedd{83.5 ± 6.7}  & \editedd{14.0 ± 8.1} &  \editedd{90.5 ± 7.9} & \editedd{21.7 ± 2.6}  \\
     & MbLS                 &  84.2 ± 4.9  & 17.9 ± 7.4 &  92.2 ± 5.6 & 26.9 ± 3.6\\
     \cdashline{2-6}
     & \editedd{TTA}        &  \editedd{83.8 ± 5.8} & \editedd{16.4 ± 7.1}  &  \editedd{92.7 ± 5.6} &  \editedd{28.8 ± 3.9} \\
     \cdashline{2-6}
     & Ensemble   &  84.5 ± 5.7 & 11.3 ± 6.5  &  94.3 ± 4.3 & 30.0 ± 4.6  \\
     \cdashline{2-6}
     & Bayes      &  84.0 ± 5.8 & \underline{8.6 ± 4.7}   &  94.7 ± 3.1 & 29.1 ± 4.8  \\
     & Bayes+AvU  &  \textbf{\underline{84.9 ± 6.9}} & 8.9 ± 6.0   &  \textbf{\underline{95.7 ± 3.2}}$^{\dag}$ & \textbf{\underline{30.5 ± 4.5}}$^{\dag}$  \\

    %%%%%%%%%%%% STRSeg %%%%%%%%%%%%
    \hline
    \multirow{11}{*}{\parbox{2.0cm}{\centering OOD \\ ------------ \\ Prostate MR \\ (PR12)}} 
     & Det        &  74.2 ± 12.6 & 15.6 ± 6.3  &  87.9 ± 7.5 & 22.1 ± 6.2 \\
     & \editedd{Det + AvU}  &  \editedd{74.5 ± 13.0} & \editedd{27.6 ± 14.3} &  \editedd{88.2 ± 7.6} & \editedd{22.0 ± 7.1} \\
     \cdashline{2-6}
     & Focal      &  71.2 ± 17.4 & 12.1 ± 5.8  &  89.0 ± 7.1 & 24.3 ± 6.7 \\
     & \editedd{ECP}        &  \editedd{74.8 ± 12.5} & \editedd{22.3 ± 10.2} &  \editedd{87.2 ± 8.1} & \editedd{20.6 ± 7.0} \\ 
     & \editedd{LS}         &  \editedd{74.5 ± 13.0} & \editedd{21.7 ± 11.5} &  \editedd{79.5 ± 8.9} & \editedd{19.1 ± 7.2} \\
     & \editedd{SVLS}       &  \editedd{76.9 ± 11.5} & \editedd{17.9 ± 9.3}  &  \editedd{87.2 ± 7.2} & \editedd{16.4 ± 5.2} \\
     & MbLS       &  73.6 ± 12.5 & 19.9 ± 7.4  &  86.5 ± 7.2 & 21.8 ± 5.6 \\
     \cdashline{2-6}
     & TTA        &  \editedd{74.0 ± 12.8} & \editedd{23.7 ± 11.4} &  \editedd{88.6 ± 7.4} & \editedd{24.9 ± 5.8}\\
     \cdashline{2-6}
     & Ensemble   &  \underline{76.3 ± 12.2} & \underline{9.7 ± 5.0}   &  \underline{91.6 ± 5.2} & \underline{28.4 ± 5.7} \\
     \cdashline{2-6}
     & Bayes      &  70.6 ± 16.6 & 11.8 ± 7.2  &  89.1 ± 7.4 & 25.7 ± 5.1 \\
     & Bayes+AvU  &  76.3 ± 12.6 & 11.4 ± 6.7  &  \textbf{90.6 ± 6.9}$^{\dag}$ & \textbf{26.2 ± 7.4}$^{\dag}$ \\
    
    \hline
    \end{tabular}
    \label{tab:results_table_pros}
\end{table}

%%%%%%%%%%%%%%%%%%%%%%%%%%%%%%%%%%%%%%%%%%%%%%% Text-HN
% Point 1 - AvU does good for Bayes (DICE + ECE + unc-err). 
% Point 2 - AvU is always better than det, calib and TTA models for unc-err and in many cases for ECE. 
% Point 3 - Ens offers competitve performance to the AvU model across datasets and metrics.  
% Point 4 - AvU loss has no effect on Det for unc-err metrics in iD, but poorer in OOD. Also, DICE is similar (in iD), but reduced in OOD.
Results in \blueautoref{tab:results_table_hn} showed that the AvU loss on the \textit{Bayes} model significantly improved calibrative and uncertainty-error correspondence (unc-err) metrics for both in-distribution (ID) and out-of-distribution (OOD) datasets. The \textit{Bayes+AvU} model also always performed better than the \textit{Det}, calibration-focused and \textit{TTA} models for unc-err metrics. Also, its ECE scores were in most cases better than calibration-focused models. However, there was no clear distinction between the performance of the \textit{Ensemble} and \textit{Bayes+AvU} model for ECE and unc-err metrics across both datasets. Also, the AvU loss did not benefit the unc-err metrics for the \textit{Det} model, in  both datasets. Of all the calibration-focused models, \textit{LS} had the lowest ECE and unc-err metrics, while the \textit{ECP} model had the best unc-err metrics. When compared to \textit{Det}, the \textit{TTA} model improved calibrative and unc-err metrics for the OOD dataset, while maintaining it for the ID dataset.

\begin{figure}[p]
    \subfloat[H\&N CT (RTOG) (in-distribution)\label{fig:results_visual_hn_rtog}]{
    	\includegraphics[width=0.97\columnwidth]{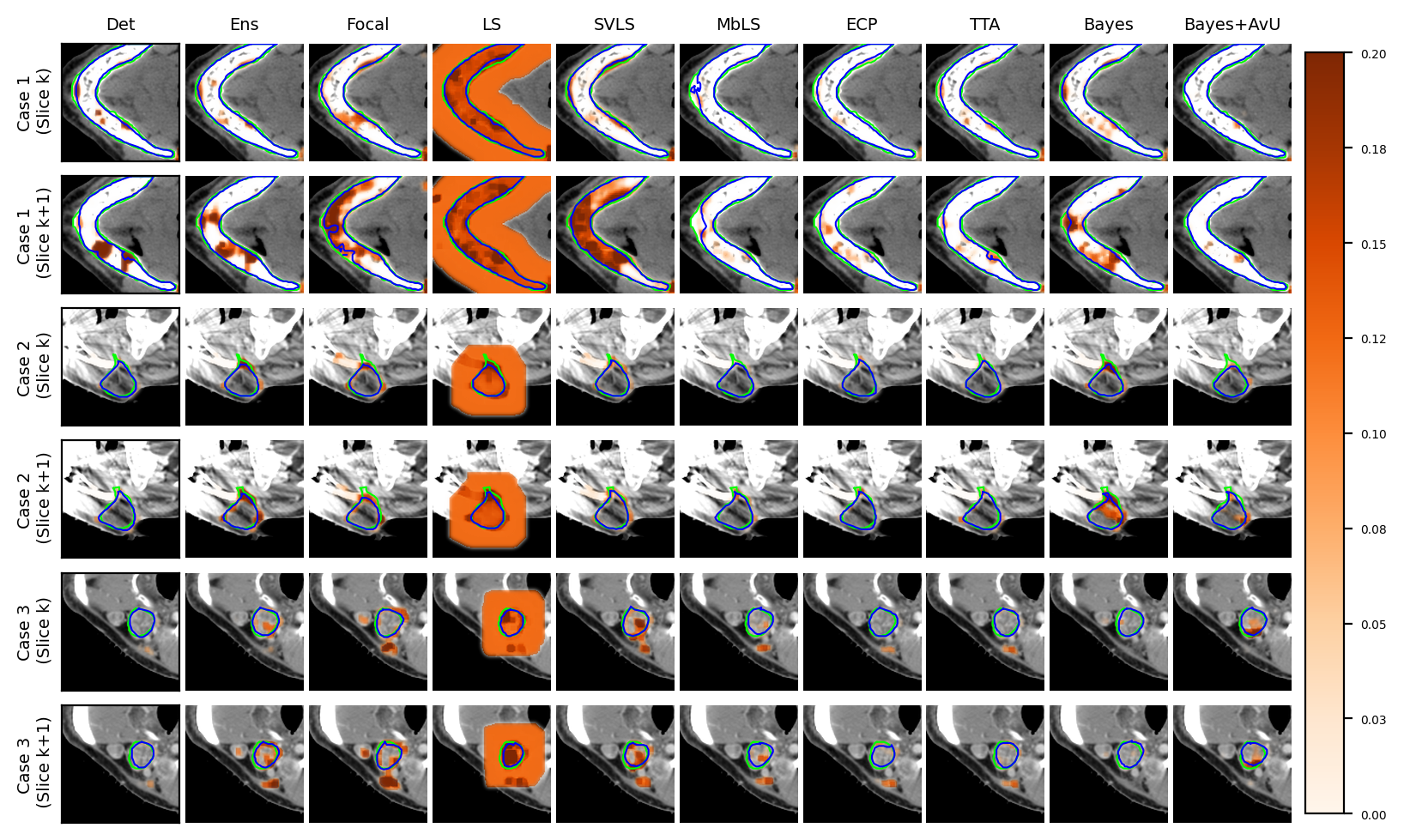}
        
    }
    \\
    \subfloat[H\&N CT (STRSeg) (out-of-distribution)\label{fig:results_visual_hn_structseg}]{
    	\includegraphics[width=0.97\columnwidth]{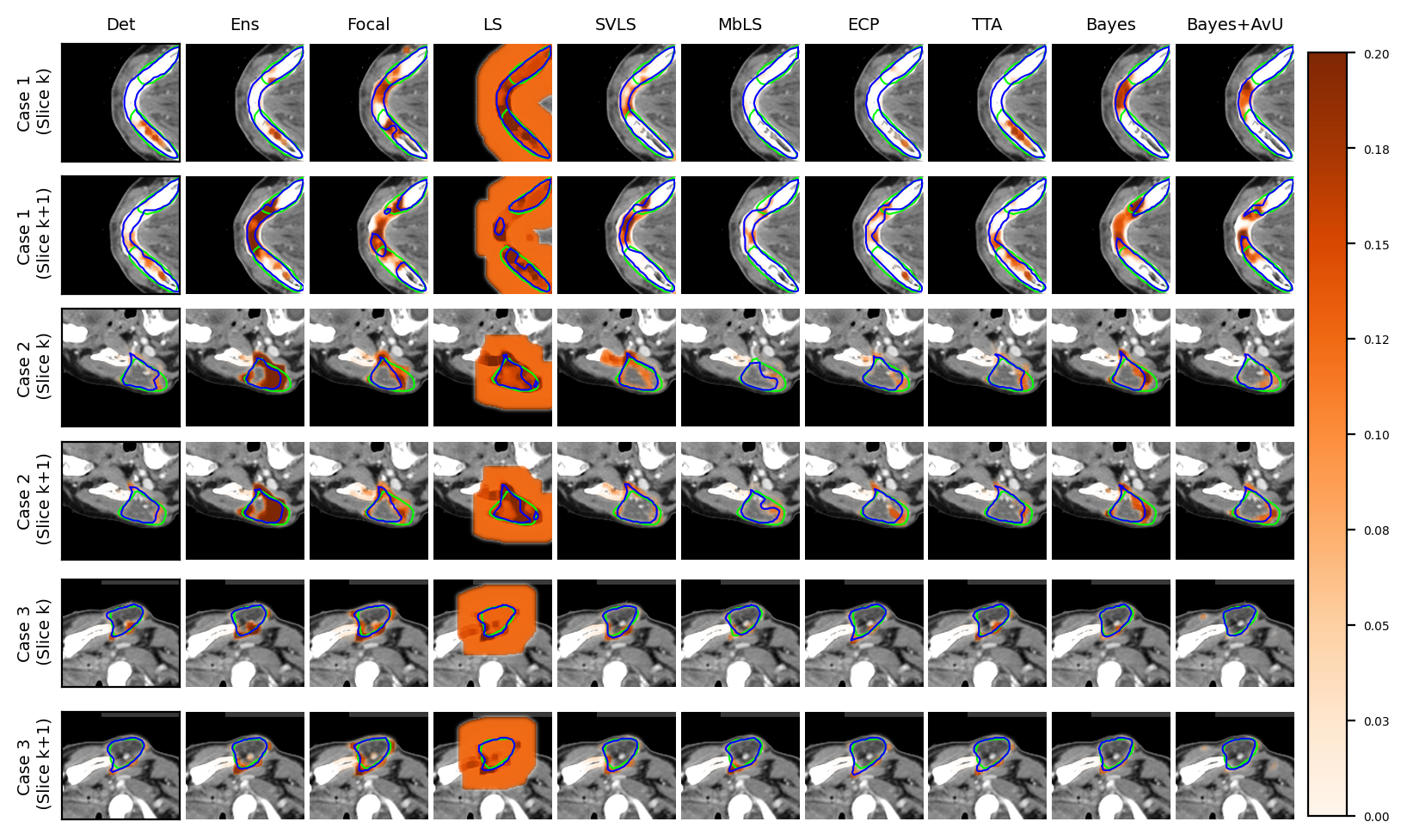}
    }
    
    \caption{Uncertainty-error correspondence for the head-and-neck (H\&N) CT (a,b) dataset. Slices of the CT scans are shown in pairs to understand the 3D nature of segmentation uncertainty heatmaps. The color bar on the right depicts the range of uncertainty values while green and blue are used for ground truth and prediction contours respectively.}
    \label{fig:results_visual_hn}
\end{figure}

Visually, the \textit{Bayes+AvU} model was able to successfully suppress uncertainty in the true positive (TP) (Case 1/2 in \blueautoref{fig:results_visual_hn_rtog}) and true negative (TN) (Case 3 in \blueautoref{fig:results_visual_hn_rtog}) regions of the predicted contour. Moreover, it also showed uncertainty in false positive (FP) regions while also suppressing uncertainty in TP regions (Case 3 in \blueautoref{fig:results_visual_hn_structseg}). Calibrative models (e.g. \textit{Focal}, \textit{LS}, \textit{SVLS}) tended to be quite uncertain in TP or TN regions, which may lead to additional QA time. Detailed descriptions are provided in \blueautoref{sec:app_visualresults}.

% Results of these models can be found in Appendix \ref{sec:app_bayesh}.

% MICCAI/PDDCA dataset [Appendix (Pg 15, Table S8) = https://jmir.org/api/download?alt_name=jmir_v23i7e26151_app1.pdf&filename=584abec9f29d69baaab930a03ecb2c2d.pdf]
% BStem=84.2, Mandible=93.8, Par(L)=88.1, Par(R)=86.6, SMD(L)=76.5. SMD(R)=79.2, Avg=84.7

% The Medical Segmentation Decathlon [Appendix (Pg 15, Table 8, nnUNet) = https://static-content.springer.com/esm/art%3A10.1038%2Fs41467-022-30695-9/MediaObjects/41467_2022_30695_MOESM1_ESM.pdf]
% nnU-Net (TZ=0.76, PZ=0.90, avg=0.83)

\subsubsection{Prostate MR}
\label{sec:results_pros}

%%%%%%%%%%%%%%%%%%%%%%%%%%%%%%%%%%%%%%%%%%%%%%% Text-Pros
% Point 1 - AvU sig improves unc-err for Bayes.  Maintains or improve DICE. Has one of the highest DICE vals. No sig improv for ECE.
% Point 2 - (similar) Better ECE than unc-err than Det, calib, TTA models. 
% Point 3 - Ens vs Bayes+AvU (similar DICE, bayes better in ID, slightly worse in OOD for unc-err & ECE).
% Point 4 - AvU loss has no noiceable effect on unc-err for iD, but poorer for OOD.
Similar to the head-and-neck CT dataset, the use of the AvU loss on the baseline \textit{Bayes} model significantly improved its uncertainty-error correspondence (unc-err) while maintaining calibration performance (\blueautoref{tab:results_table_pros}). Moreover, it improved the DICE values such that its one of the most competitive amongst all models. Also, the \textit{Bayes+AvU} had better performance in both unc-err and calibrative metrics when compared to the \textit{Det}, calibration-focused and \textit{TTA} models. When comparing to the \textit{Ensemble}, the \textit{Bayes+AvU} had similar DICE. While \textit{Bayes+AvU} had better calibrative and unc-err performance in the in-distribution (ID) dataset, the \textit{Ensemble} performed better in the out-of-distribution (OOD) setting. The AvU loss had no positive effect on the DICE and unc-err performance of the \textit{Det} model in both the ID and OOD setting, however there was an increase in ECE.

\begin{figure}[p]
    \subfloat[Prostate MR (PrMedDec) (in-distribution)\label{fig:results_visual_pros_prmeddec}]{
    	\includegraphics[width=0.97\columnwidth]{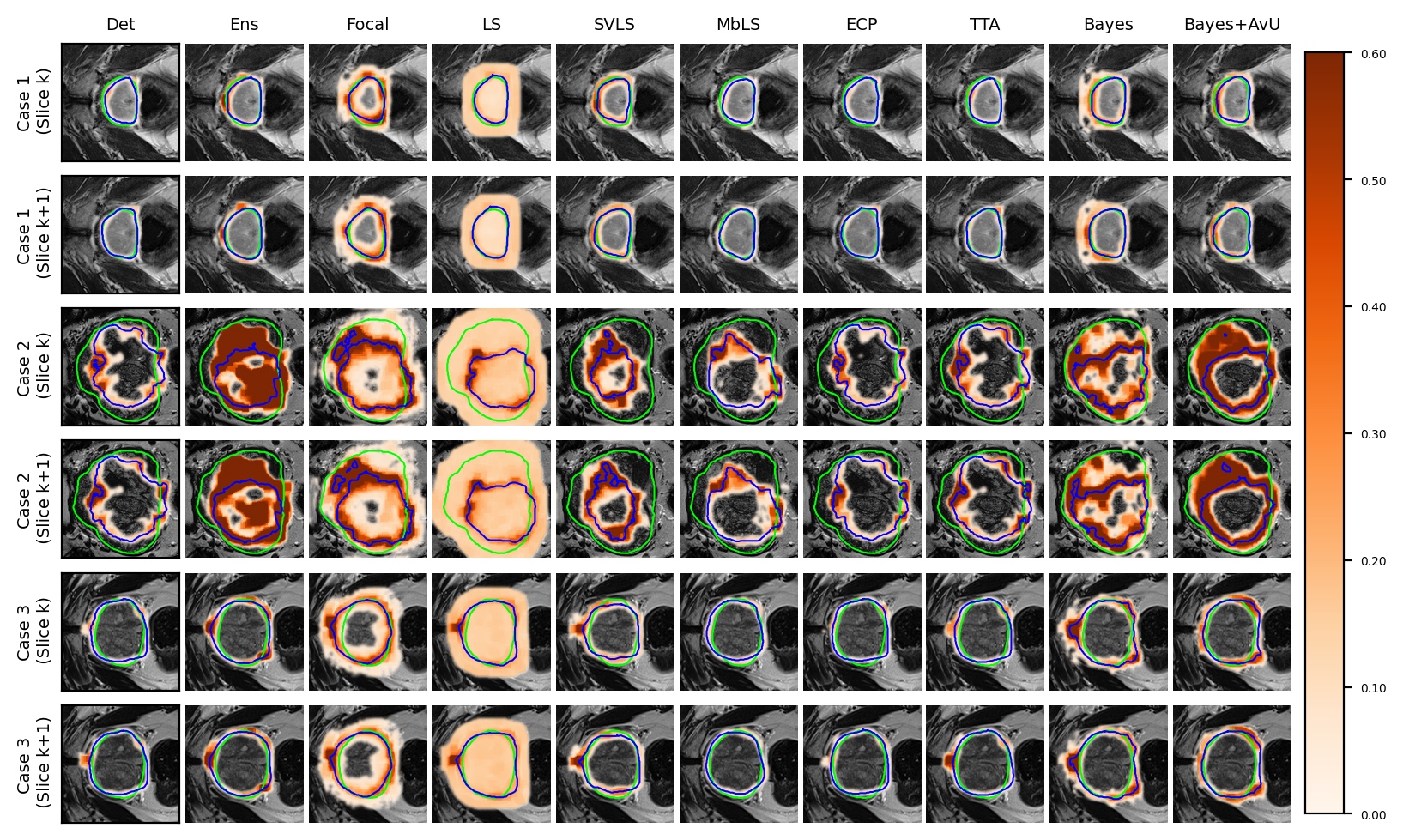}
        
    }
    \\
    \subfloat[Prostate MR (PR12) (out-of-distribution)\label{fig:results_visual_pros_pr12}]{
    	\includegraphics[width=0.97\columnwidth]{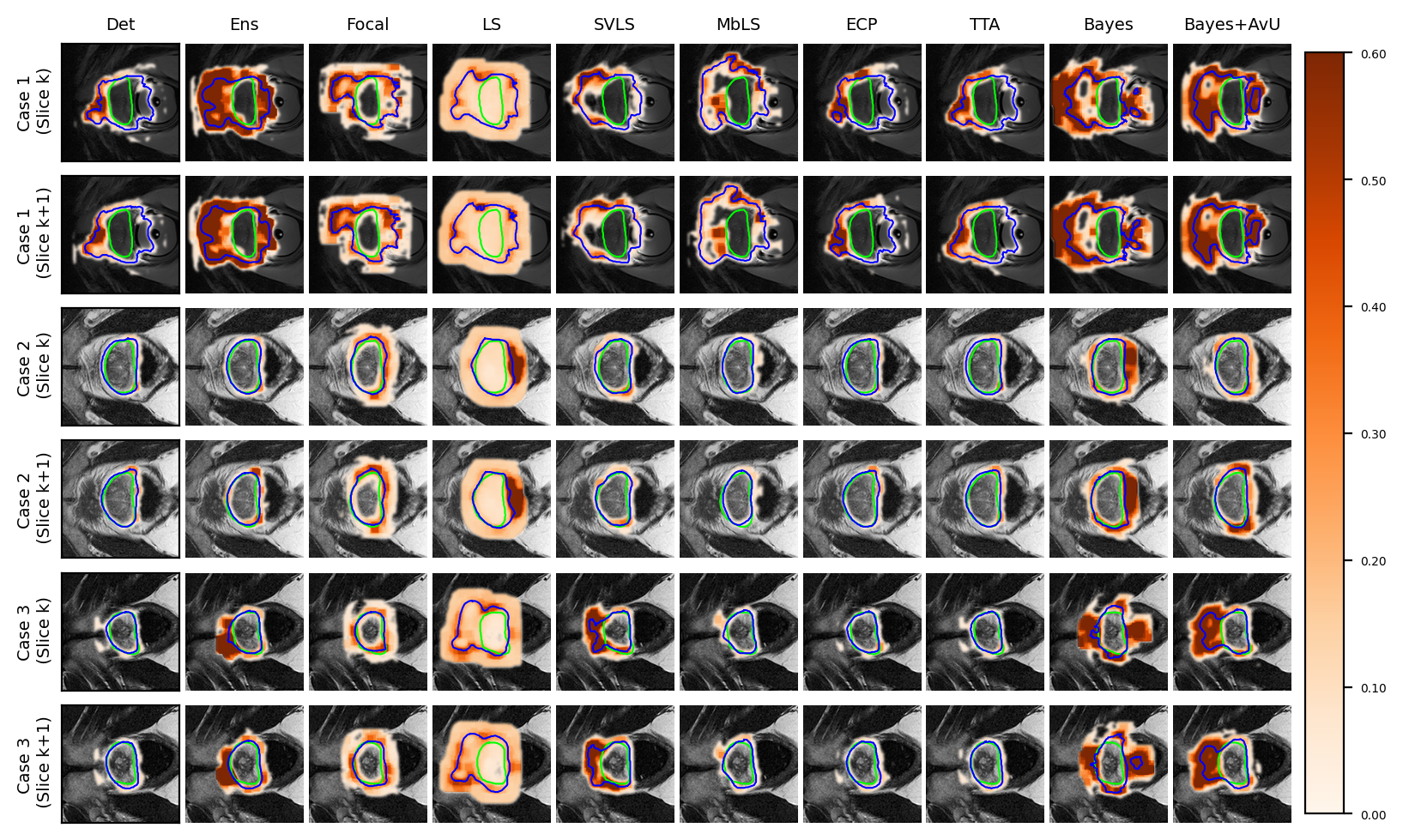}
    }
    \caption{Uncertainty-error correspondence for the Prostate MR (a,b) dataset. Slices of the MR scans are shown in pairs to understand the 3D nature of segmentation uncertainty heatmaps. The color bar on the right depicts the range of uncertainty values while green and blue are used for ground truth and prediction contours respectively.}
    \label{fig:results_visual_pros}
\end{figure}

Visual results show that the \textit{Bayes+AvU} successfully suppresses uncertainty in the true negative (Case 1 in \blueautoref{fig:results_visual_pros_prmeddec}, Case 2 in \blueautoref{fig:results_visual_pros_pr12}) and true positive (Case 2 in \blueautoref{fig:results_visual_pros_prmeddec}) regions of the predicted contour. It also shows uncertainty in the false positive regions (Case 2 in \blueautoref{fig:results_visual_pros_prmeddec}, Case 1/3 in \blueautoref{fig:results_visual_pros_pr12})  

%%%%%%%%%%%%%%%%%%%%%%%%%%%%%%%%%%%%%%%%%%%%%%% Figure-HN (unc on CT)
% \setkeys{Gin}{draft=false}
% \setkeys{Gin}{draft}

%%%%%%%%%%%%%%%%%%%%%%%%%%%%%%%%%%%%%%%%%%%%%%% Figure-Pros (unc on MR)
% \setkeys{Gin}{draft=false}
% \setkeys{Gin}{draft}

%%%%%%%%%%%%%%%%%%%%%%%%%%%%%%%%%%%%%%%%%%%%%%%% Figure (plots)
% \setkeys{Gin}{draft=false}
\begin{figure}[!t]

    %%%%%%% Curves
    \subfloat[AvU Curve (RTOG)]{
    	\includegraphics[width=0.32\textwidth]{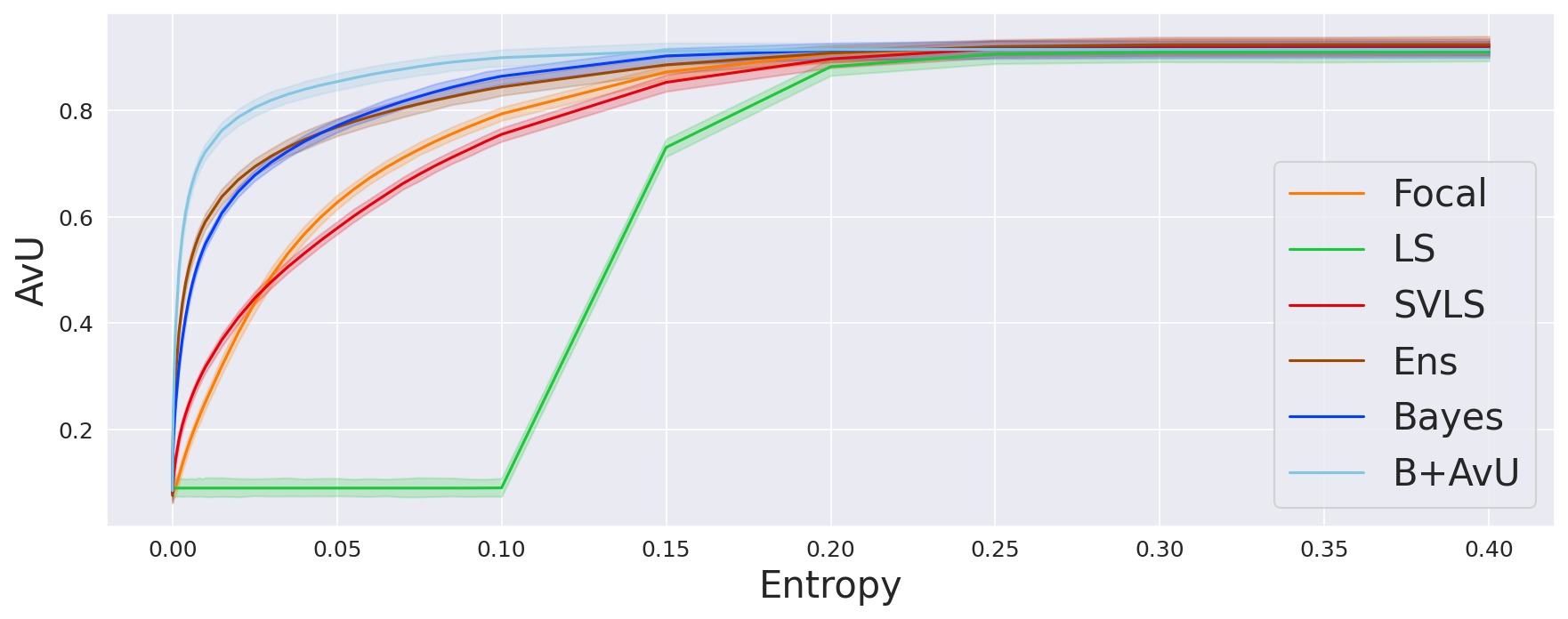}
    }
    \subfloat[ROC Curve (RTOG)]{
    	\includegraphics[width=0.32\textwidth]{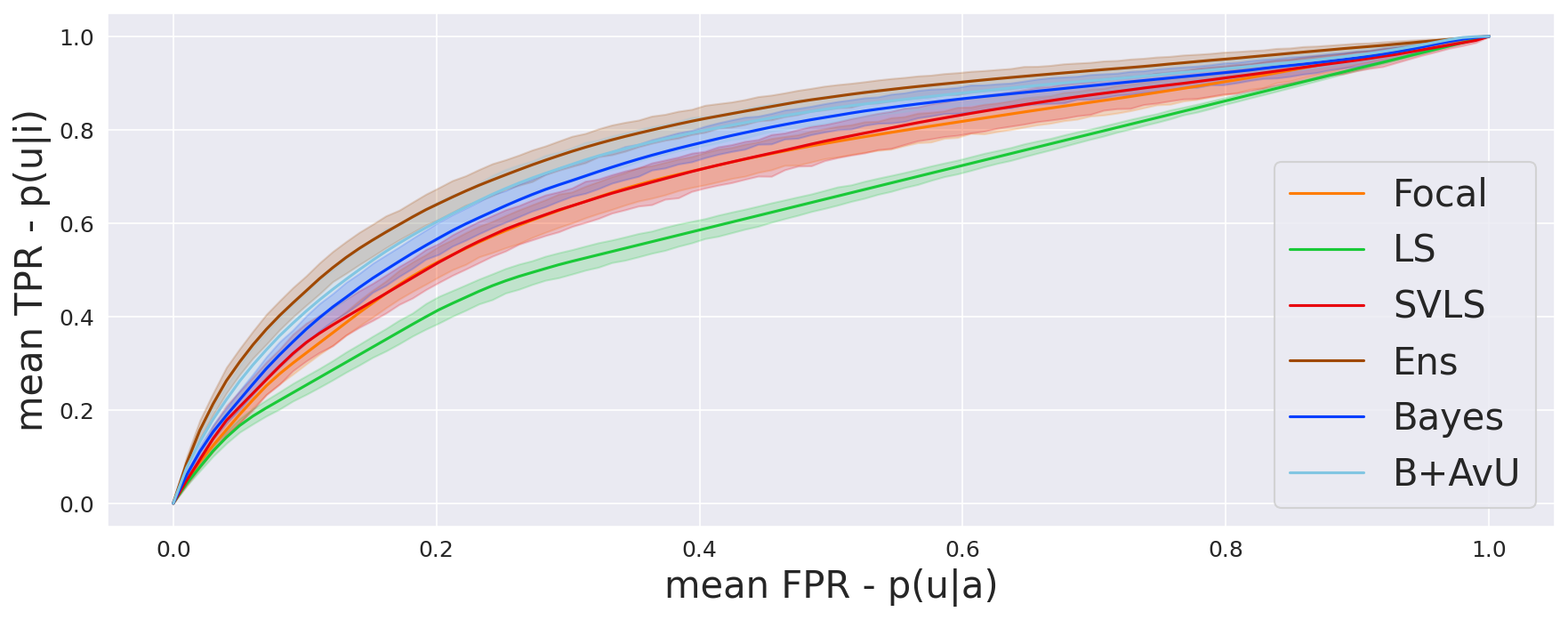}
    }
    \subfloat[PRC Curve (RTOG)]{
    	\includegraphics[width=0.32\textwidth]{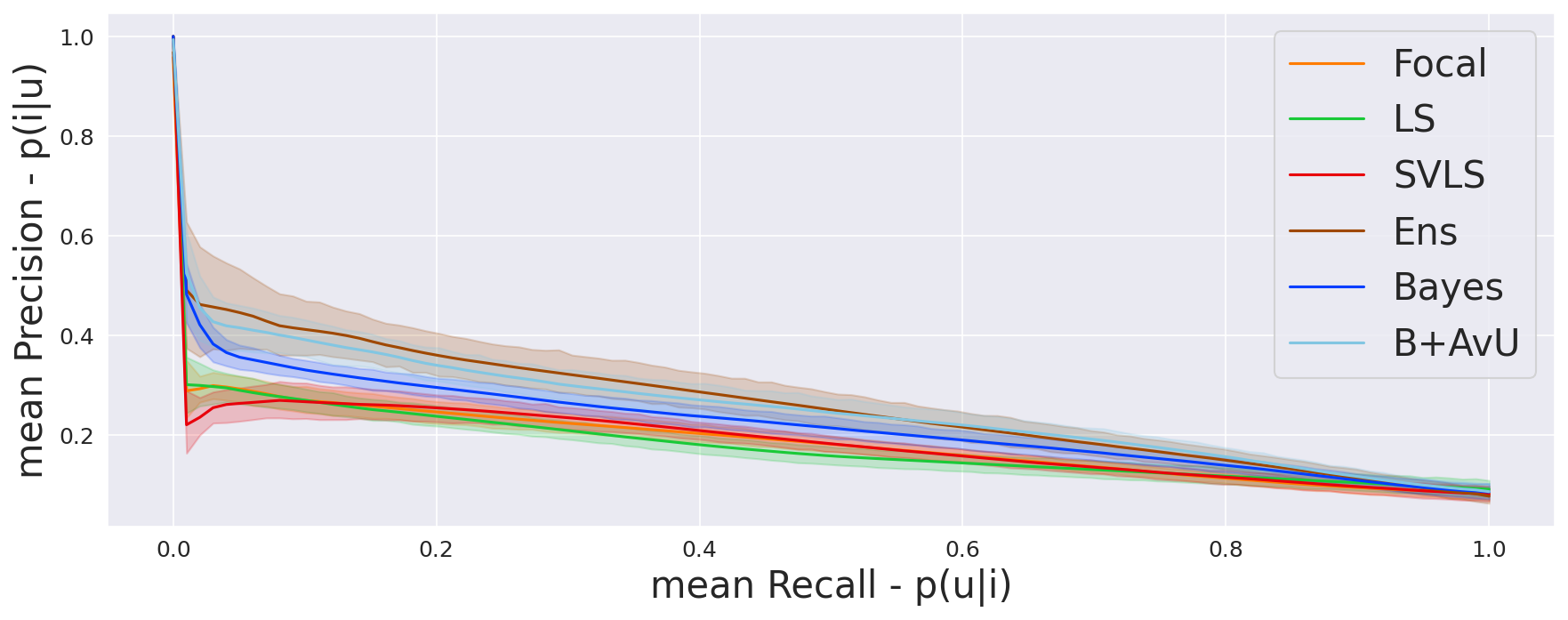}
    }

    %%%%%%% BoxPlots
    \subfloat[AvU Boxplot (RTOG)]{
    	\includegraphics[width=0.32\textwidth]{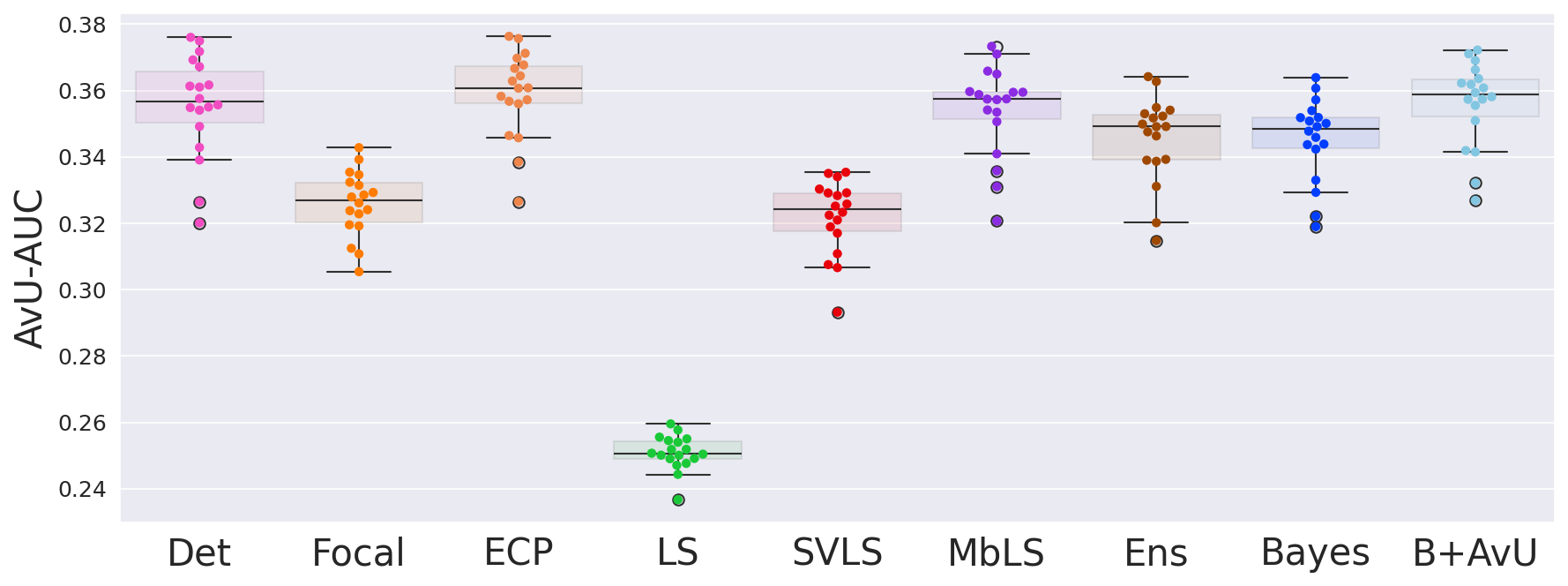}
    }
    \subfloat[ROC Boxplot (RTOG)]{
    	\includegraphics[width=0.32\textwidth]{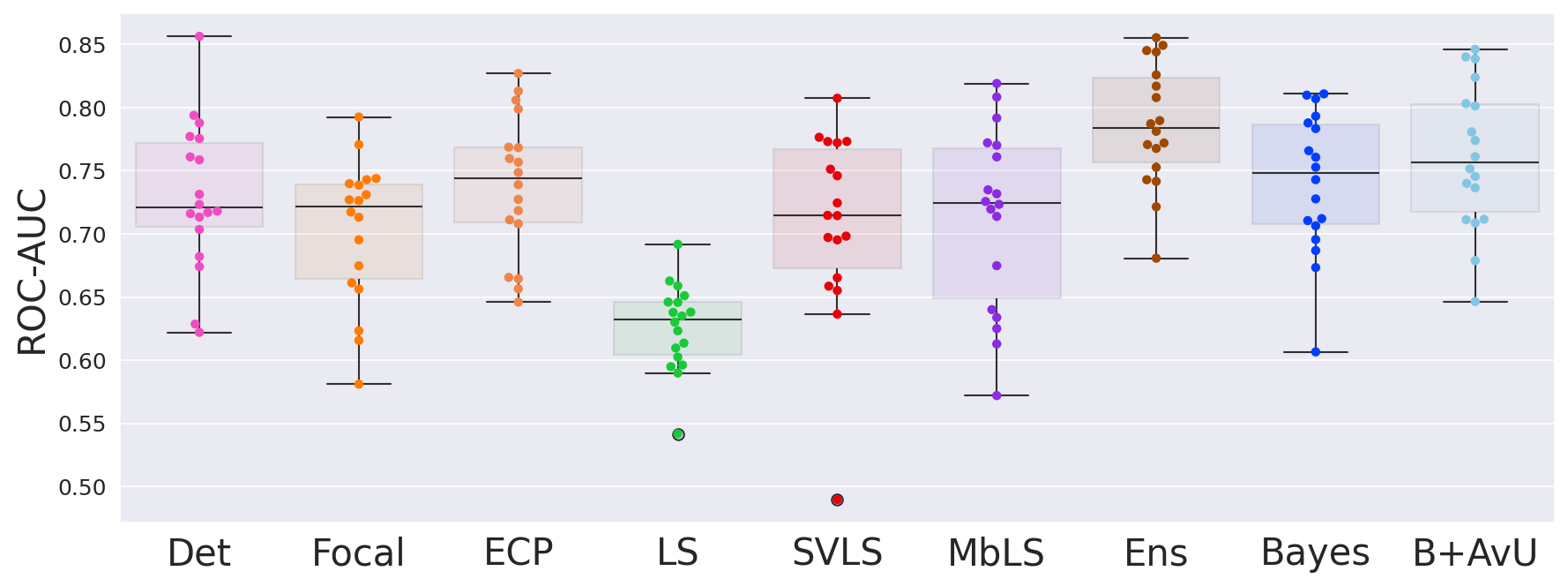}
    }
    \subfloat[PRC Boxplot (RTOG)]{
    	\includegraphics[width=0.32\textwidth]{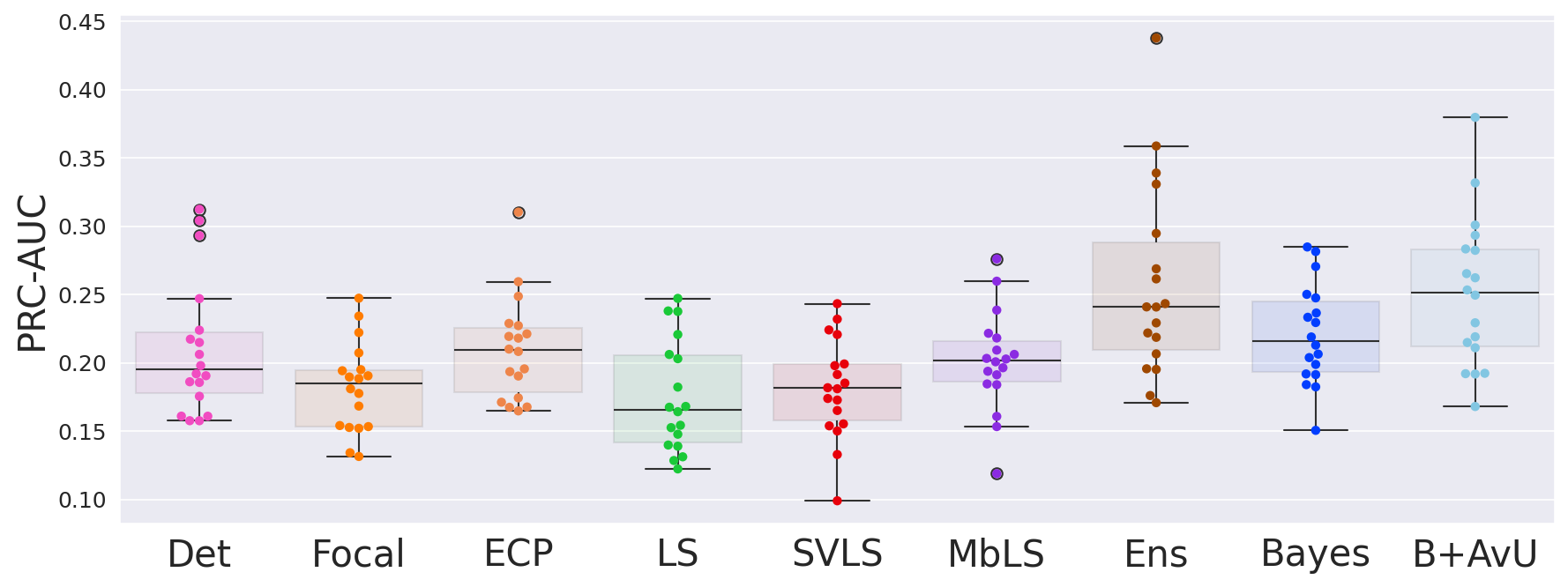}
    }

    %%%%%%% Curves
    \subfloat[AvU Curve (MedDec)]{
    	\includegraphics[width=0.32\textwidth]{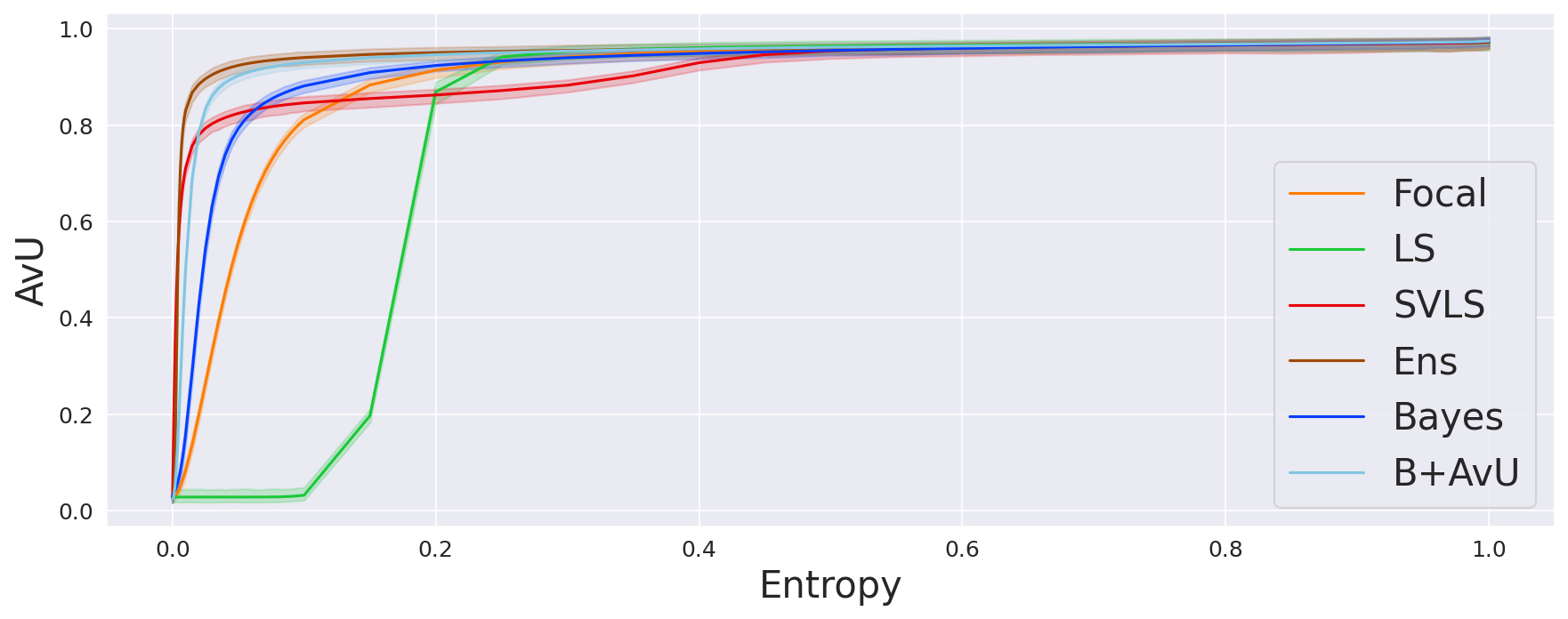}
    }
    \subfloat[ROC Curve (PrMedDec)]{
    	\includegraphics[width=0.32\textwidth]{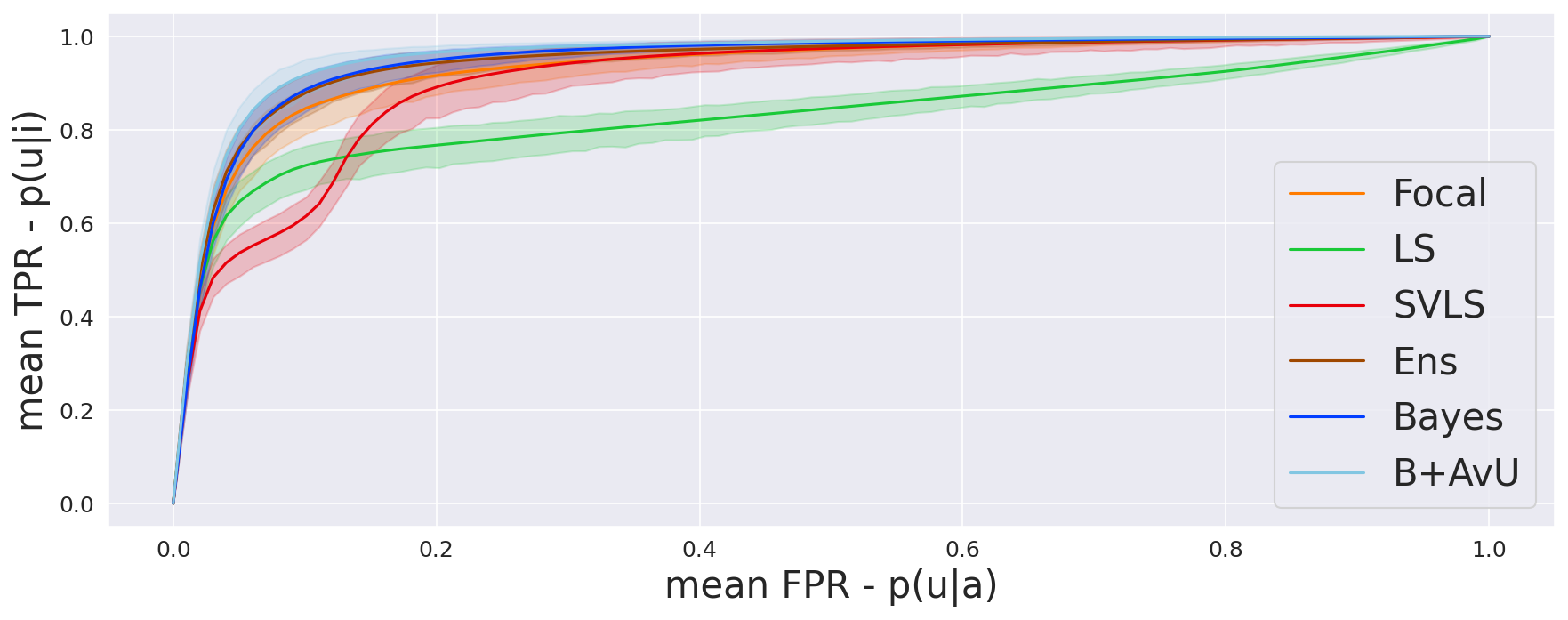}
    }
    \subfloat[PRC Curve (PrMedDec)]{
    	\includegraphics[width=0.32\textwidth]{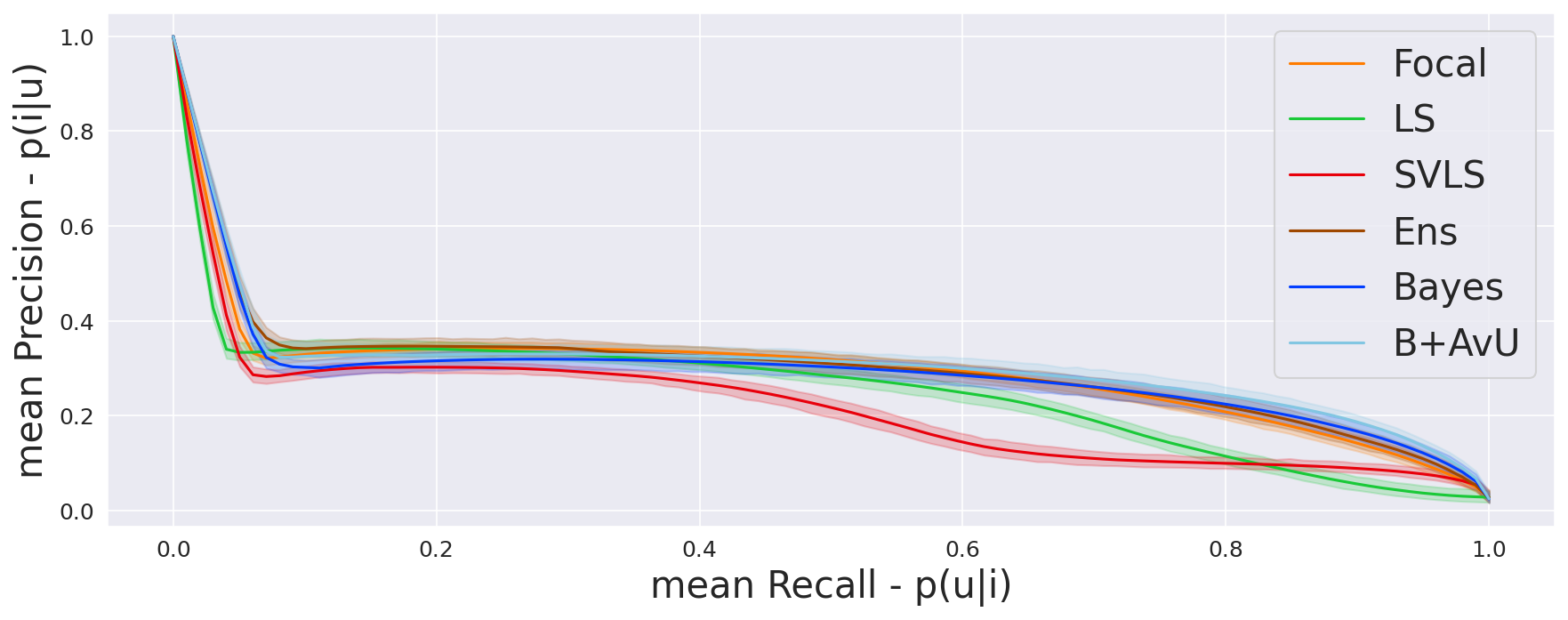}
    }

    %%%%%%% BoxPlots
    \subfloat[AvU Boxplot (PrMedDec)]{
    	\includegraphics[width=0.32\textwidth]{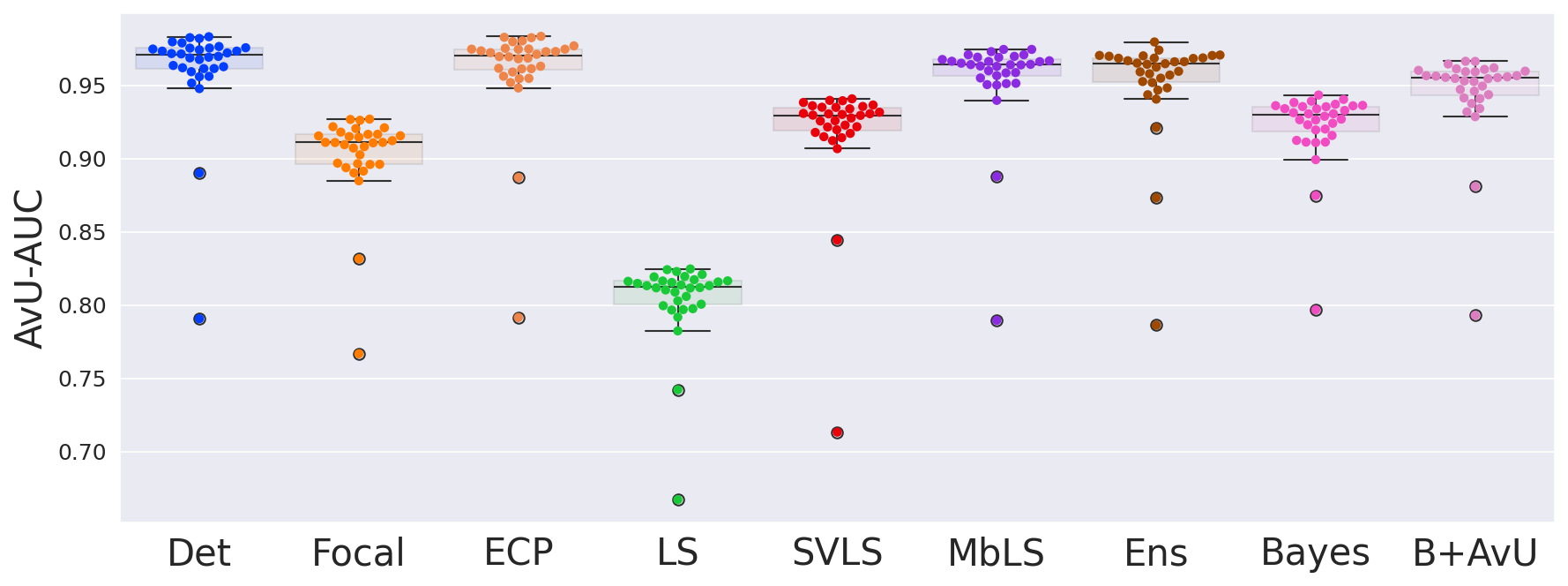}
    }
    \subfloat[ROC Boxplot (PrMedDec)]{
    	\includegraphics[width=0.32\textwidth]{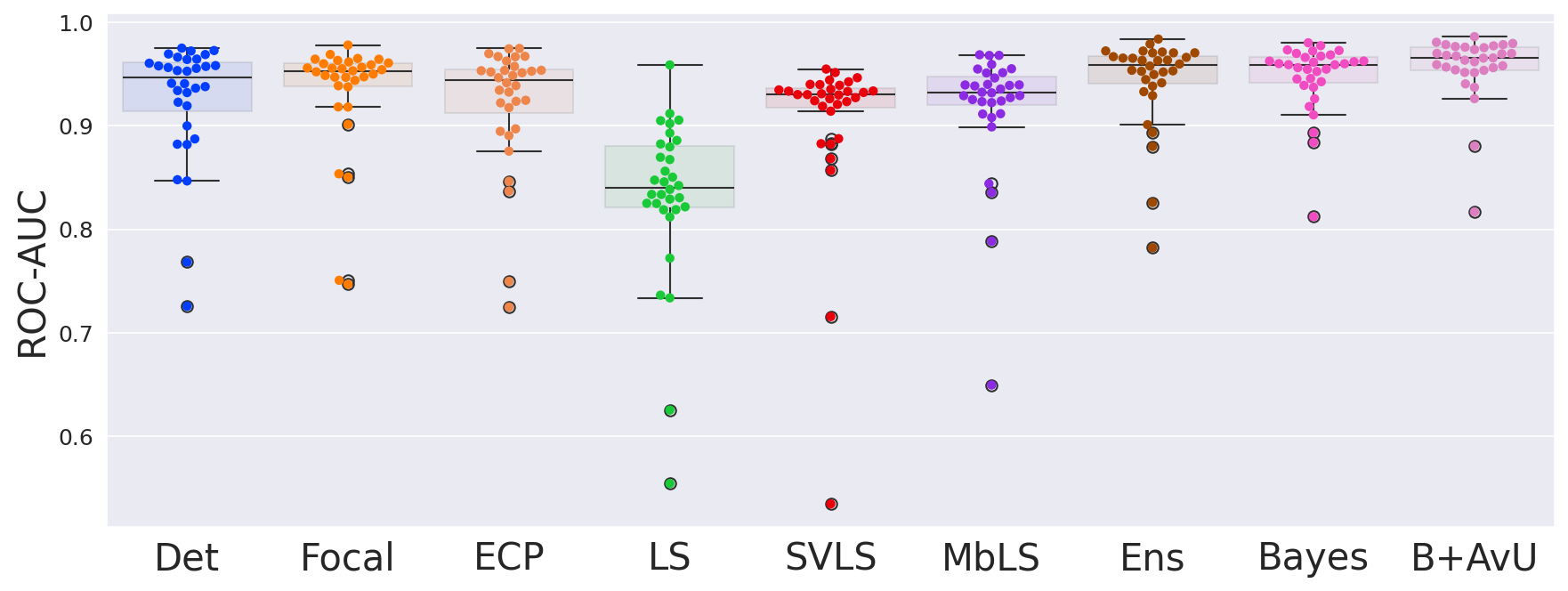}
    }
    \subfloat[PRC Boxplot (PrMedDec)]{
    	\includegraphics[width=0.32\textwidth]{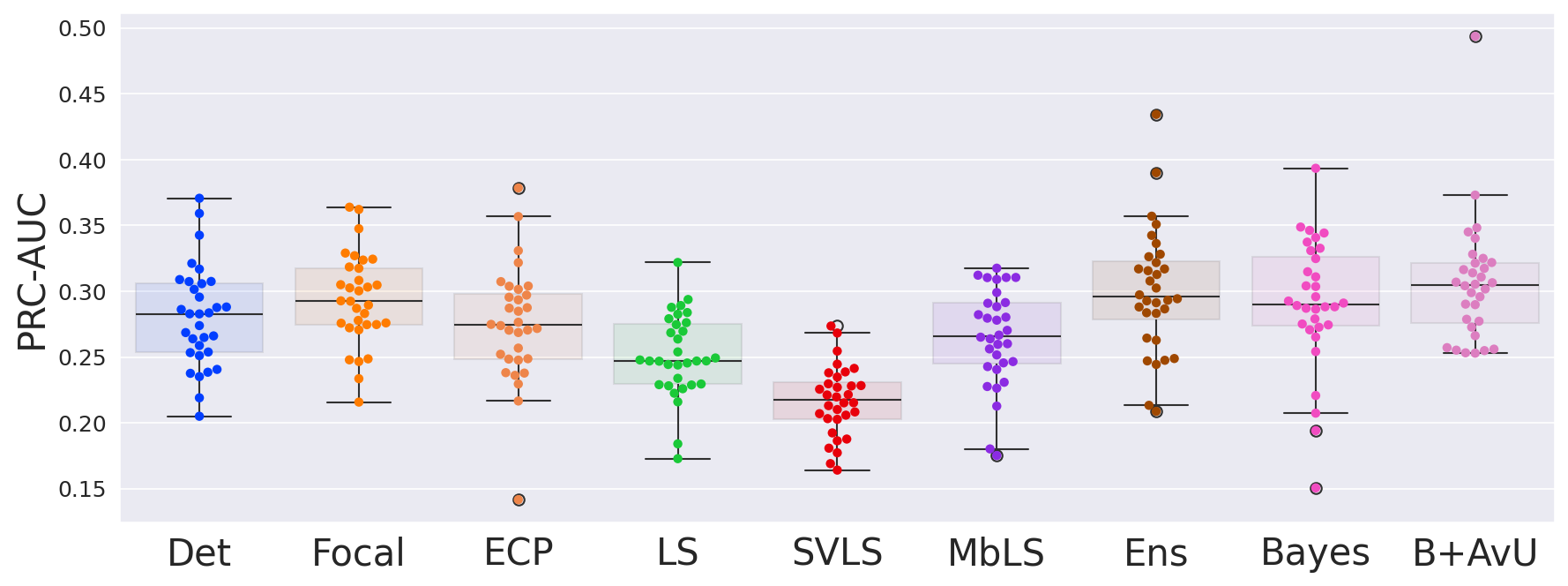}
    }
\caption{The figures above show the distribution of the uncertainty-error correspondence metrics as curves and boxplots (with swarm plots) for patients from the RTOG clinical trial (a-f) as well as for the Medical Decathlon (Prostate) dataset (g-l). We only evaluate up to the maximum uncertainty of each dataset as the metrics do not change beyond that.}
\label{fig:results_curves}
\end{figure}

% P =[0.75, 0.25] = 0.75*ln(0.75) + 0.25*ln(0.25) = -0.562/ln(2) = -0.8107 (normalized uncertainty)
% P = [0.75, 0.04, 0.04, 0.04, 0.04, 0.04, 0.04] = (0.75*ln(0.75) + 6*0.04*ln(0.04))/ln(7) = -0.507
\section{Discussion}
Although medical image segmentation using deep learning can now predict high quality contours which can be considered clinically acceptable, a manual quality assessment (QA) step is still required in a clinical setting. To truly make these models an integral part of clinical workflows, we need them to be able to express their uncertainty and for those uncertainties to be useful in a QA setting. To this end, we test 11 models which are either Bayesian, deterministic, calibration-focused or ensembled.

% Point 1 - Generic discussion

\subsection{Discriminative and Calibrative Performance}
In context of DICE and ECE, the use of the AvU loss on the baseline \textit{Bayes} model always showed results which have never statistically deteriorated. Moreover, the DICE results for the in-distribution (ID) head-and-neck dataset (RTOG) were on-par with existing state-of-the-art models (83.6 vs 84.7 for \cite{dataset_deepmind}). The same held for the ID Prostate dataset (PRMedDec) where results were better than advanced models (84.9 vs 83.0 for \cite{dataset_meddec}). These results validate the use of our neural architecture \citep{model_organnet2.5d}, and training strategy.  

Secondly, although the \textit{Ensemble} model, in general, had better or equivalent DICE and ECE scores across all 4 datasets, it also required 3x more parameters than the \textit{Bayes+AvU} model. Also, as expected, and due to 5x more parameters, the \textit{Ensemble} model performed better than the \textit{Det} model for DICE and ECE. 

Finally, in the regime of segmentation ``failures" as the inaccuracy map, the calibrative methods did not generally have improved calibration performance when compared to the \textit{Det} model. In theory, these models regularize the model's probabilities by making it more uncertain and hence avoid overconfidence. In practice however, this leads to the predicted contours being uncertain along their accurate boundaries, most evident in visual examples of the \textit{Focal} and \textit{SVLS} model (see \blueautoref{fig:results_visual_hn} and \blueautoref{fig:results_visual_pros}). Also, visual image characteristics in different regions of the scan that are similar to the segmented organs may cause these models to showcase uncertainty in those areas (for e.g. patches of uncertainty in Case 3 of \blueautoref{fig:results_visual_hn_rtog}).

% import numpy as np
% softmax = lambda x: np.exp(x) / np.sum(np.exp(x), axis=0)

% softmax([-5, 5]) = [~0, ~1]
% softmax([-5, 5, 1, 1, 1, 1]) = [~0, 0.93, 0.017, 0.017, 0.017, 0.017]

\subsection{Uncertainty-Error Correspondence Performance}
Although calibrative metrics are useful to compare the average truthfulness of a model's probabilities, they may not be relevant to real-world usage in a pixel-wise segmentation QA scenario. Considering a clinical workflow in which uncertainty can be used as a proxy for error-detection, we evaluate the correspondence between them. Results showed that across both in- and out-of-distribution datasets, the \textit{Bayes+AvU} model has one of the highest uncertainty-error correspondence metrics. Similar trends were observed for the \textit{BayesH+AvU} (\blueautoref{sec:app_bayesh}) model, however \textit{Bayes+AvU} was better. We hypothesize that this is due to perturbations in the bottleneck of UNet-like models having a better understanding of semantic concepts (e.g., shape, size etc) than the decoder layers. However, the AvU loss did not offer benefit to the \textit{Det} model on both datasets indicating that this loss may rely on the model to already exhibit some level of uncertainty.

An interesting case is shown in \blueautoref{fig:results_visual_hn_structseg} (Case 3) which showed uncertainty on the white blob (a vein) in the middle of the grey tissue of the organ. Many models showed uncertainty on the vein due to a difference in its texture from that of the organ. However, this information may be distracting to a clinician as they are using uncertainty for error detection. Given that there were no segmentation \textquotedblleft failures\textquotedblright, our \textit{Bayes+AvU} model successfully suppressed all uncertainties. In another case (\blueautoref{fig:results_visual_hn_rtog} - Case 3), we saw that for 3D segmentation, uncertainty is also 3D in nature. Our \textit{Bayes+AvU} model had an error in the second slice and correctly showed uncertainty there. However, this uncertainty overflowed on the first slice and hence penalized the uncertainty-error correspondence metrics. Such results indicate that during contour QA, the clinician can \editeddd{potentially} trust our AvU loss models more than other models as they are better indicative of potential errors. This reduces time wasted analyzing false positive regions (i.e., accurate but uncertain) and hence increases trust between an expert and deep learning-based contour QA tools. Also note that in general, the two-class prostate dataset visually showcased higher levels of uncertainty than the six-class head-and-neck dataset. 
% - more distraction, more time needed, less trust
% -(0.8*ln(0.8) + 0.2*ln(0.2))/ln(2)     = 0.721
% -(0.8*ln(0.8) + 5*0.04*ln(0.04))/ln(6) = 0.458

As seen in \blueautoref{tab:results_table_hn}, \blueautoref{tab:results_table_pros} and \blueautoref{fig:results_curves}, there is no clear choice between the top two performing models i.e., \textit{Bayes+AvU} and \textit{Ensemble} for uncertainty-error correspondence. The visual results, however, indicate that the \textit{Ensemble} model is more uncertain in accurate regions. Also, for all the datasets, the \textit{Det} model has high AvU scores when compared to the \textit{Bayes+AvU} model (\blueautoref{sec:appendix_c}). Here, it is important to consider that the AvU metric (\blueautoref{eq:6_avu_terms}) is essentially uncertainty accuracy, and thus, also comes with its own pitfalls. Given that all models had a DICE value which leads to more accurate terms and less inaccurate terms, the AvU metric got skewed due to the large count of $n_\mathrm{ac}$ terms. However, upon factoring the ROC and PRC curves, it becomes evident that the \textit{Det} model is not the best performing for uncertainty-error correspondence. 

Finally, all calibration-focused methods - \textit{Focal, ECP, LS, SVLS} and \textit{MBLS} had ROC and PRC metrics lower than the baseline \textit{Bayes} model indicating that training for model calibration may not necessarily translate to uncertainty outputs useful for error detection. 

\subsection{Future Work}
In a radiotherapy setting, the goal is to maximize radiation to tumorous regions and minimize it for healthy organs. This goal is often not optimally achieved due to imperfect contours caused by time constraints and amorphous region-of-interest boundaries on medical scans. Thus, an extension of our work could evaluate the contouring corrections made by clinicians in response to uncertainty-proposed errors in context of the dose changes to the different regions of interest. Such an experiment can better evaluate the clinical utility of an uncertainty-driven error correction workflow. 
% Blobby-error analysis (Fig 4b - Case 2)

\section{Conclusion}
This work investigates the usage of the Accuracy-vs-Uncertainty (AvU) metric to improve clinical ``utility" of deep Bayesian uncertainty as a proxy for error detection in segmentation settings. Experimental results indicate that using a differentiable AvU metric as an objective to train Bayesian segmentation models has a positive effect on uncertainty-error correspondence metrics. We show that our AvU-trained Bayesian models have equivalent or improved uncertainty-error correspondence metrics when compared to various calibrative and uncertainty-based methods. Given that our approach is a loss function, it can be used with other neural architectures capable of estimating uncertainty.  
% Our approach is neural architecture agnostic, and also applicable to popular models like nnUnet \citep{isensee2021nnu}.

Given that deep learning models have shown the capability of reaching near expert-level performance in medical image segmentation, one of the next steps in their evolution is evaluating their clinical utility. Our work shows progress on this using a uncertainty-driven loss in a Bayesian setting. We do this for two radiotherapy body-sites and modalities as well in an out-of-distribution setting. Our hope is that the community is inspired by our positive results to further contribute to human-centric approaches to deep learning-based modeling.

% Pros: Since ROC-AUC of Ensemble is same as that of Bayes and Bayes+AvU, we take a look at PRC-AUC of Ensemble. Since PRC-AUC of Ensemble is lower than that of Bayes and Bayes+AvU, it means that at the same level of recall (x-axis), precision of Ensemble is lower. Since precision contains the False Positive (n_au) term, it means that Ensemble shows high amounts of uncertainty in accurate regions, which is also what we notice in the images. 

%%%%%%%%%%%%%%%%%%%%%%%%%%%%%%%%%%%%%%%%%%%%%%%%%%%%%%%%%%%%%%%%%%%%%%%
% Mandatory Sections. Please complete, especially for final publication
%%%%%%%%%%%%%%%%%%%%%%%%%%%%%%%%%%%%%%%%%%%%%%%%%%%%%%%%%%%%%%%%%%%%%%%

% Acknowledgements.
% Please include any funding, intellectual contributions not included in the authorship, and any other acknowledgements.
\acks{The research for this work was funded by Varian, a Siemens Healthineers Company, through the HollandPTC-Varian Consortium (grant id 2019022) and partly financed by the Surcharge for Top Consortia for Knowledge and Innovation (TKIs) from the Ministry of Economic Affairs and Climate, The Netherlands.}

% Ethical Standards.
% Please edit with the appropriate ethics considerations for your work. Include any pertinent IRB information, etc.
%
% Please note that the submission requirements included:
% The work presented must follow appropriate ethical standards in conducting research and writing the manuscript, following all applicable laws and regulations regarding treatment of animals or human subjects.
\ethics{The work follows appropriate ethical standards in conducting research and writing the manuscript, following all applicable laws and regulations regarding treatment of animals or human subjects.}

% Conflict of Interest
% Declaration of possible conflicts of interest: Authors must disclose any financial, organisational, commercial or personal conflicts of interest that might bias their work.
% If no conflicts, please say "We declare we don't have conflicts of interest."
\coi{We declare we don't have conflicts of interest.}

\bibliography{main-melba}

% needed in general before appendices.
% \newpage

% Appendix is optional
\clearpage
\appendices

\section{Segmentation ``Failures" and ``Errors"}
\label{sec:appendix_a}

\setkeys{Gin}{draft=false}
% Figure (seg error vs failure)
\begin{figure}[!tbh]
    \subfloat[Contours\label{fig:results_visual_inacc_contours}]{
    	\includegraphics[width=0.252\columnwidth,frame]{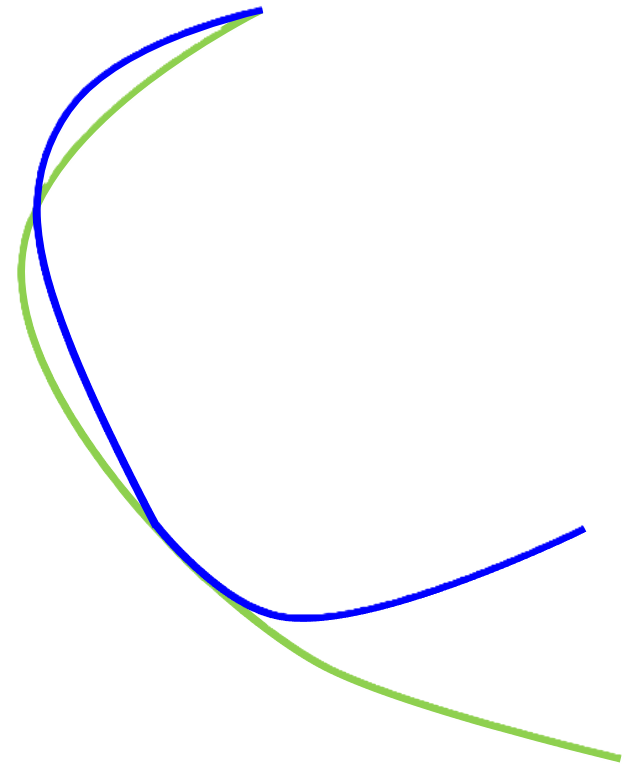}
    }
    \subfloat[Inaccuracy Map\label{fig:results_visual_inacc_inaccmap}]{
    	\includegraphics[width=0.23\columnwidth,frame]{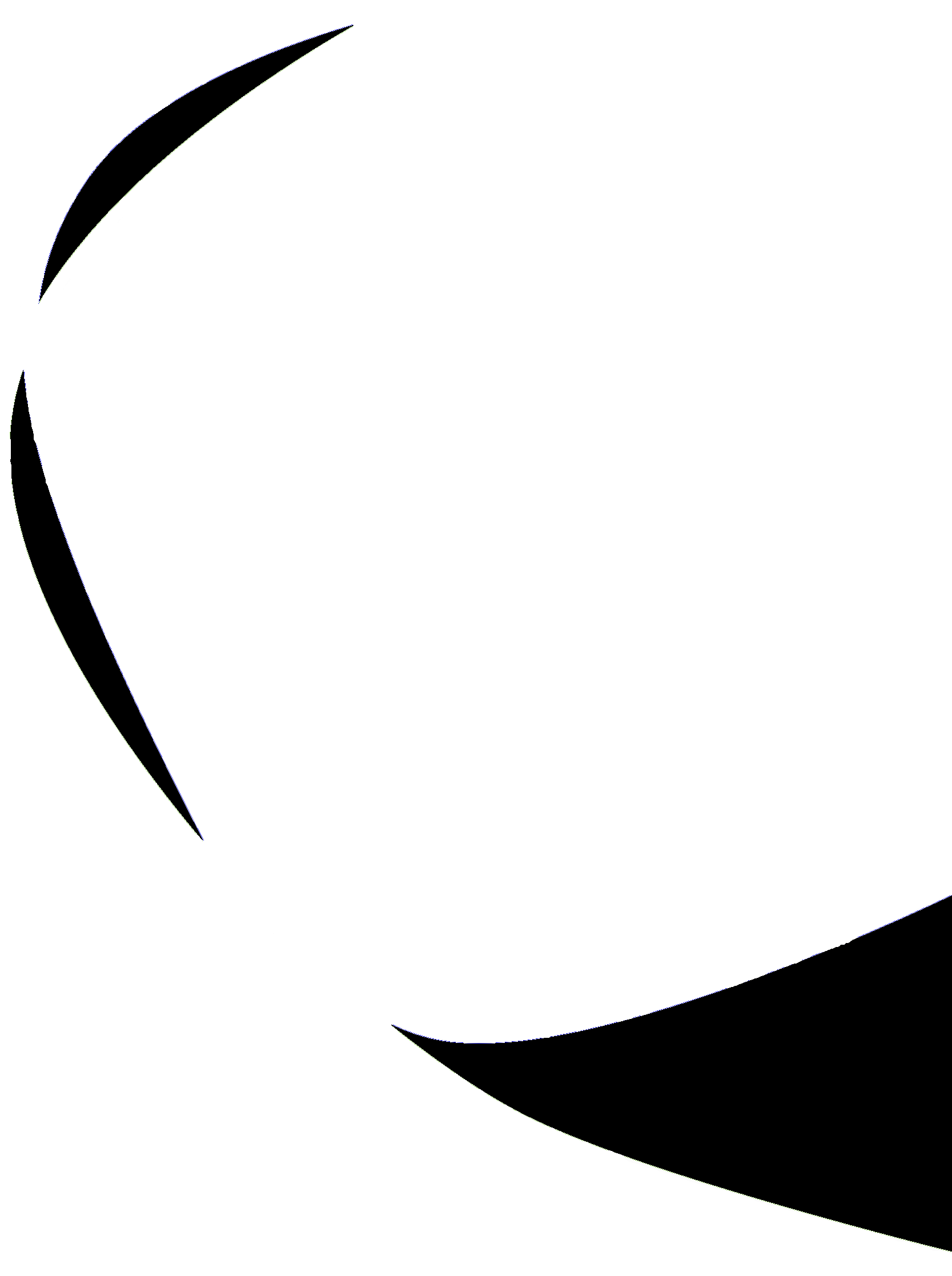}
    }
    \subfloat[Segmentation Errors\label{fig:results_visual_inacc_segerrors}]{
    	\includegraphics[width=0.23\columnwidth,frame]{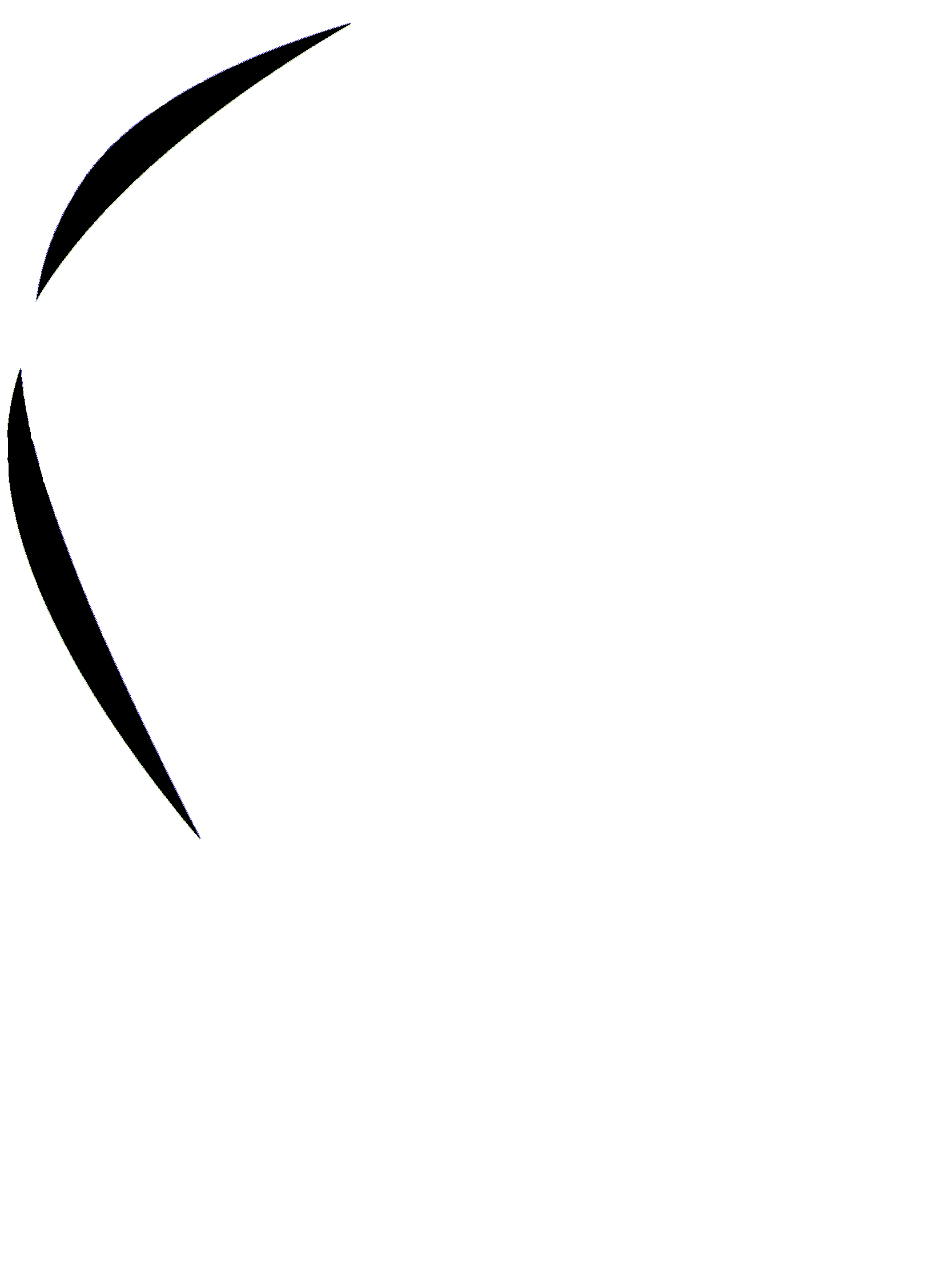}
    }
    \subfloat[Segmentation Failures\label{fig:results_visual_inacc_segfailures}]{
    	\includegraphics[width=0.23\columnwidth,frame]{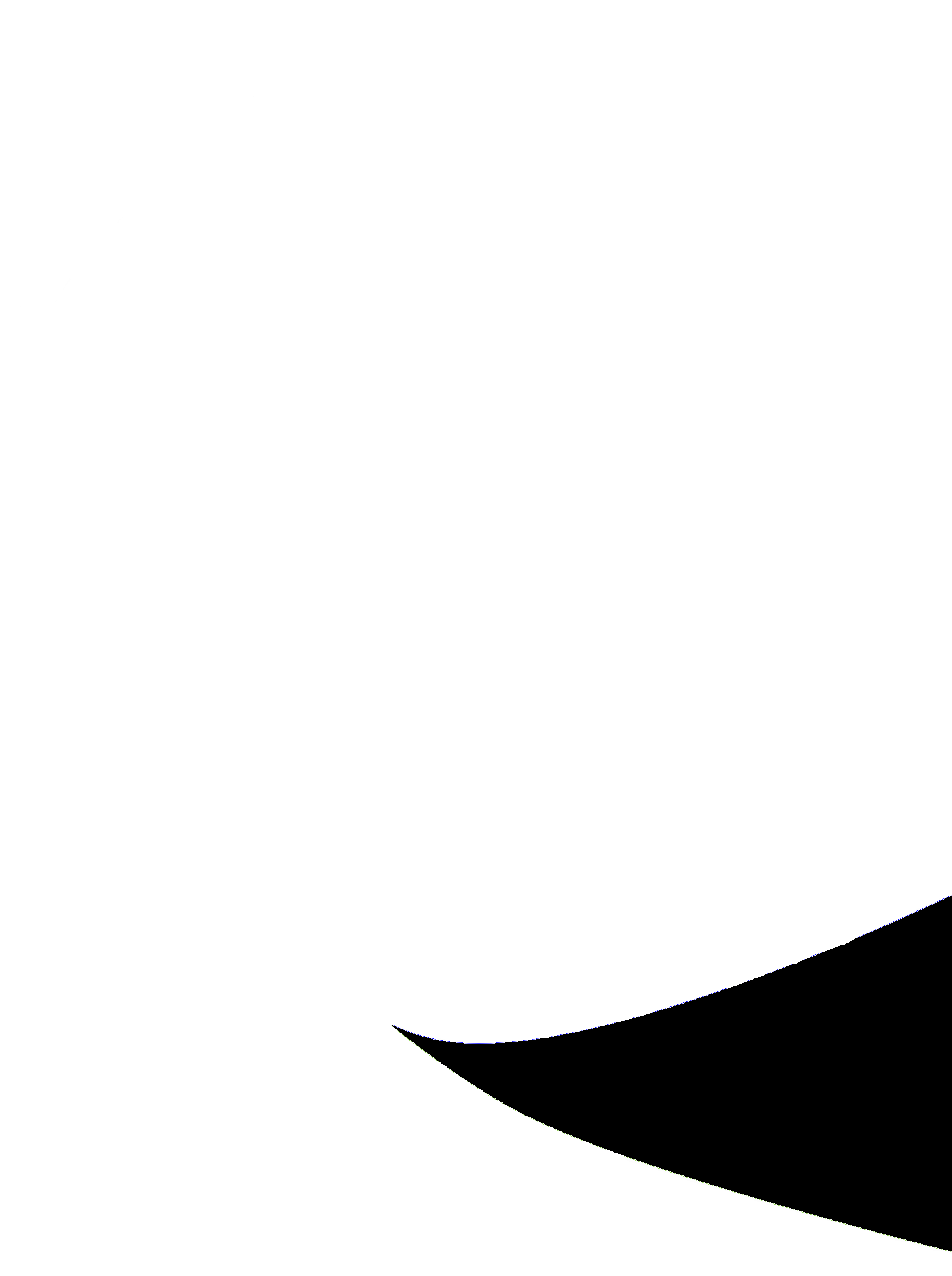}
    }
    
\caption{The green and blue contours in a) show the ground truth (GT) and predicted contours. In b) we see the inaccuracy map in black, while c) and d) show the smaller segmentation \textquotedblleft errors\textquotedblright\ and larger segmentation \textquotedblleft failures\textquotedblright\ respectively.}
\label{fig:results_visual_inacc}
\end{figure}

\section{Weightage of AvU loss}
\label{app_validation_avu}
The table below show the weights used for the AvU loss which were finetuned on the validation datasets of the head-and-neck CT and prostate MR. The final weightage was chosen by identifying the inflection point at which the \textit{ROC-AUC} and \textit{PRC-AUC} drop precipitously. Given that the AvU loss is a log term, its values are inherently small ($\leq$ 1.0). This is then added to the cross-entropy term, which is a sum of logs (Eqn (3)) over all the voxels (=N) and all the classes (=C). Thus, we used a balancing term in the range of 10$^1$ to 10$^4$.

\begin{table}[h!]
    \centering
    \caption{Uncertainty-error correspondence results (higher is better) to select the weightage of the AvU loss. \underline{Underlined} numbers indicate the maximum value for a metric.}
    \begin{tabular}{|c|l|l|l|l|}

    \hline
    \thead{Validation \\ Dataset} & \thead{Model} & \thead{AvU-AUC \\(x$10^{-2}$)} & \thead{ROC-AUC \\(x$10^{-2}$)} & \thead{PRC-AUC \\(x$10^{-2}$)} \\
    \hline
    
    % Ensemble       & 33.7±1.2 & 81.4±5.2 & 31.2±6.8 \\
    \multirow{4}{*}{\parbox{3.0cm}{\centering H\&N CT \\ (MICCAI2015)}} & Bayes & 34.1 ± 0.7 & 79.1 ± 4.7 & 25.9 ± 2.9 \\
    % Bayes +1AvU    & 34.3±0.9 & 78.4±5.8 & 27.2±3.4 \\
     & Bayes + 10AvU   & 34.5 ± 0.9 & 78.2 ± 6.0 & 26.1 ± 3.4 \\
     & Bayes + 100AvU  & 35.5 ± 0.6 & \underline{79.6 ± 4.8} & \underline{28.0 ± 3.5} \\
     & Bayes + 1000AvU & \underline{35.9 ± 6.9} & 76.4 ± 5.8 & 23.1 ± 1.7 \\ 
    \hline
    \multirow{4}{*}{\parbox{3.0cm}{\centering Prostate\\MR \\ (ProstateX)}} 
    & Bayes            & 93.2 ± 1.8 & 95.3 ± 1.9 & 30.3 ± 2.9\\
    & Bayes + 100AvU   & 94.9 ± 2.1 & 95.9 ± 2.0 & 31.5 ± 3.5\\
    & Bayes + 1000AvU  & 95.5 ± 1.9 & \underline{96.3 ± 2.4} & \underline{32.0 ± 3.3}\\
    & Bayes + 10000AvU & \underline{96.1 ± 1.7} & 93.1 ± 2.1 & 29.3 ± 3.1\\ 
    \hline
    
    \end{tabular}
    
    \label{tab:results_validation}
\end{table}

\newpage

\newpage
\section{Hyperparameter selection}
\label{sec:appendix_c}

In the tables shown below, we report results for different hyperparameters of different model classes. If the DICE of a hyperparameter is 10.0 points lower than the class maximum, we ignore it. We also ignore models with large drops in ECE or AvU-AUC when compared to models in its own class. To choose the best hyperparameter, it has to perform as the best in four out of the five metrics, else we chose the middlemost hyperparameter.
% Note that the further you go down the rows for each class, the more different the model becomes compared to the deterministic model.

% Table (unc-error)
% AvU = n_ac + n_iu / (N=(n_ac + n_au + n_iu + n_ic))
% ROC = p(u|i) vs p(u|a)
% PRC = p(u|i) vs p(i|u)

%%%%%%%%%%%%%%%%%%%%%%%%%%%%%%%%%%%%
% Head-and-Neck
%%%%%%%%%%%%%%%%%%%%%%%%%%%%%%%%%%%%
\newcommand{\CA}{1}
\newcommand{\setCOL}{6}
\begin{table}[H]
\centering
\caption{Volumetric (DICE), calibrative (ECE) and uncertainty-error correspondence metrics (AvU-AUC, ROC-AUC, PRC-AUC) on head-and-neck validation dataset for the purpose of hyperparameter selection. The experiment indicated as \textbf{bold} is the one with the best performance.}
\begin{tabular}{|l|c|c|c|c|c|}
    \hline
    \thead{\textbf{Experiment}} & \thead{DICE $\uparrow$ \\(x$10^{-2}$)}  & \thead{ECE $\downarrow$ \\(x$10^{-2}$)}  & \thead{AvU-AUC $\uparrow$ \\(x$10^{-2}$)} & \thead{ROC-AUC $\uparrow$\\(x$10^{-2}$)} & \thead{PRC-AUC $\uparrow$\\(x$10^{-2}$)} \\ 
    \hline
    Det                    & 83.6 ± 2.2    & 8.5 ± 1.6  & 35.1 ± 1.1 & 74.8 ± 5.0 & 24.5 ± 0.9  \\
    Det + 10AvU            & 83.4 ± 1.7    & 8.4 ± 1.6  & 35.4 ± 1.0 & 74.4 ± 4.8 & 23.2 ± 1.4  \\
    \textbf{Det + 100AvU}  & 83.6 ± 1.4    & 8.1 ± 1.5  & 36.1 ± 0.6 & 75.6 ± 2.9 & 23.4 ± 2.1  \\
    Det + 1000AvU          & 58.1 ± 6.9    & 14.7 ± 4.0 & 30.2 ± 1.6 & 78.8 ± 4.6 & 23.0 ± 10.6 \\
    \cdashline{\CA-\setCOL}
    \textbf{Focal($\gamma$=1)} & 84.1 ± 0.8  & 8.0 ± 0.6  & 32.4 ± 0.7 & 73.9 ± 4.1 & 21.5 ± 3.4 \\ 
    Focal($\gamma$=2)          & 83.4 ± 1.3  & 9.6 ± 8.3  & 24.8 ± 1.1 & 73.7 ± 1.6 & 22.7 ± 4.1 \\ 
    Focal($\gamma$=3)          & 84.1 ± 1.9  & 15.5 ± 1.9 & 17.5 ± 7.4 & 73.1 ± 3.2 & 12.6 ± 1.9 \\ 
    \cdashline{\CA-\setCOL}
    ECP($\lambda$=0.1)           & 83.9 ± 1.3   & 8.3 ± 1.3  & 35.3 ± 0.8 & 75.1 ± 4.3 & 22.3 ± 0.8  \\ 
    \textbf{ECP($\lambda$=1.0)}  & 84.0 ± 1.1   & 8.5 ± 0.8  & 35.4 ± 0.7 & 75.3 ± 3.2 & 23.4 ± 1.4 \\ 
    ECP($\lambda$=10.0)          & 83.2 ± 2.1   & 8.7 ± 1.3  & 35.2 ± 0.8 & 74.9 ± 3.9 & 24.6 ± 2.1 \\ 
    ECP($\lambda$=100.0)         & 81.2 ± 6.4   & 17.9 ± 1.5 & 28.7 ± 2.8 & 65.4 ± 5.6 & 17.5 ± 5.2  \\ 
    \cdashline{\CA-\setCOL}
    LS($\alpha$=0.01)          & 83.0 ± 2.1  & 8.1 ± 1.3 & 32.6 ± 0.9 & 70.9 ± 2.6 & 23.4 ± 2.7  \\
    \textbf{LS($\alpha$=0.05)} & 83.6 ± 1.2  & 6.1 ± 1.0 & 24.9 ± 0.4 & 64.5 ± 3.3 & 18.1 ± 2.0  \\
    LS($\alpha$=0.1)           & 83.5 ± 1.2  & 7.9 ± 1.2 & 17.5 ± 0.1 & 63.9 ± 2.2 & 22.2 ± 1.1  \\     
    \cdashline{\CA-\setCOL}
    \textbf{SVLS($\sigma$=1)}  & 83.5 ± 1.3 & 7.7 ± 0.7 & 32.3 ± 0.8 & 71.5 ± 2.5 & 19.9 ± 0.4  \\ 
    SVLS($\sigma$=2)           & 83.5 ± 1.7 & 8.1 ± 0.9 & 31.8 ± 1.0 & 70.5 ± 3.8 & 17.7 ± 1.5  \\ 
    SVLS($\sigma$=3)           & 84.1 ± 2.0 & 7.7 ± 0.7 & 31.9 ± 1.0 & 71.3 ± 4.4 & 19.2 ± 3.2  \\ 
    \cdashline{\CA-\setCOL}
    MbLS($\lambda=0.1$,m=30)          & 82.7 ± 1.8    & 8.5 ± 0.6    & 34.9 ± 1.0 & 74.0 ± 4.3 & 23.1 ± 1.2 \\
    \textbf{MbLS($\lambda=0.1$,m=20)} & 84.4 ± 1.4    & 8.0 ± 1.1    & 35.2 ± 0.7 & 72.3 ± 3.3 & 20.4 ± 1.0 \\
    MbLS($\lambda=0.1$,m=10)          & 82.7 ± 1.8    & 8.5 ± 0.6    & 32.9 ± 0.7 & 68.4 ± 3.0 & 21.7 ± 2.2 \\
    MbLS($\lambda=0.1$,m=8)           & 62.9 ± 7.6    & 18.75 ± 1.4  & 26.0 ± 0.4 & 74.9 ± 4.0 & 39.1 ± 2.7 \\
    \hdashline[0.5pt/5pt]
    MbLS($\lambda=1$,m=20)            & 83.2 ± 1.3    & 8.9 ± 1.5    & 35.0 ± 0.9 & 72.4 ± 4.4 & 22.5 ± 1.1 \\
    \textbf{MbLS($\lambda=10$,m=20)}  & 83.4 ± 1.4    & 8.5 ± 2.0    & 34.2 ± 1.1 & 72.2 ± 4.4 & 23.1 ± 2.0 \\
    MbLS($\lambda=100$,m=20)          & 81.8 ± 1.8    & 8.0 ± 1.1    & 32.1 ± 0.9 & 69.6 ± 4.8 & 21.0 ± 2.3  \\
    \cdashline{\CA-\setCOL}
    TTA                         & 83.5 ± 2.2    & 8.5 ± 1.7  & 34.9 ± 1.1 & 75.3 ± 5.2 & 25.2 ± 1.7 \\ 
    \cdashline{\CA-\setCOL}
    Ens          & 84.9 ± 1.6    & 6.8 ± 0.9 & 34.1 ± 1.1 & 80.8 ± 3.2 & 28.2 ± 4.1  \\ 
    \cdashline{\CA-\setCOL}
    Bayes                     & 84.2 ± 2.9  & 7.8 ± 1.3 & 34.1 ± 0.7 & 79.1 ± 4.7 & 25.9 ± 2.9  \\
    Bayes + 10AvU             & 83.1 ± 2.9  & 7.6 ± 2.0 & 34.5 ± 0.9 & 78.2 ± 6.0 & 26.1 ± 3.4  \\
    \textbf{Bayes + 100AvU}   & 83.2 ± 1.7  & 7.0 ± 1.9 & 35.5 ± 0.6 & 79.6 ± 4.8 & 28.0 ± 3.5  \\
    Bayes + 1000AvU           & 84.3 ± 1.0  & 7.5 ± 1.5 & 35.9 ± 6.9 & 76.4 ± 5.8 & 23.1 ± 1.7  \\
     \hline
\end{tabular}
\end{table}

\begin{table}[H]
\centering
\caption{Volumetric (DICE), calibrative (ECE) and uncertainty-error correspondence metrics (AvU-AUC, ROC-AUC, PRC-AUC) on head-and-neck iD dataset. The experiment indicated as \textbf{bold} is the one with the best performance. * indicates hyperparameters chosen by the validation dataset.}
\begin{tabular}{|l|c|c|c|c|c|}
    \hline
    \thead{\textbf{Experiment}} & \thead{DICE $\uparrow$ \\(x$10^{-2}$)}  & \thead{ECE $\downarrow$ \\(x$10^{-2}$)}  & \thead{AvU-AUC $\uparrow$ \\(x$10^{-2}$)} & \thead{ROC-AUC $\uparrow$\\(x$10^{-2}$)} & \thead{PRC-AUC $\uparrow$\\(x$10^{-2}$)} \\ 
    \hline
    Det                     & 84.2 ± 2.7  & 9.0 ± 2.1  & 35.5 ± 1.5 & 73.0 ± 5.7 & 21.0 ± 4.8  \\
    Det + 10AvU             & 83.7 ± 2.3  & 9.3 ± 2.2  & 35.7 ± 1.3 & 70.6 ± 5.3 & 20.0 ± 3.6   \\
    \textbf{Det + 100AvU}*  & 83.8 ± 2.9  & 8.6 ± 2.7  & 36.2 ± 1.4 & 73.1 ± 6.0 & 20.8 ± 4.0    \\
    Det + 1000AvU           & 62.3 ± 5.6  & 12.1 ± 2.9 & 30.7 ± 1.2 & 78.0 ± 4.6 & 16.0 ± 9.0  \\
    \cdashline{\CA-\setCOL}
    \textbf{Focal($\gamma$=1)}* & 84.3 ± 2.4 & 9.3 ± 1.5  & 32.5 ± 0.9 & 70.3 ± 5.5 & 18.2 ± 3.2  \\ 
    Focal($\gamma$=2)           & 84.2 ± 2.0 & 11.2 ± 1.6 & 25.1 ± 0.7 & 69.4 ± 4.9 & 17.2 ± 3.0  \\ 
    Focal($\gamma$=3)           & 83.9 ± 2.5 & 15.7 ± 2.3 & 17.9 ± 5.3 & 70.5 ± 5.0 & 12.2 ± 2.9  \\ 
    \cdashline{\CA-\setCOL}
    ECP($\lambda$=0.1)           & 84.4 ± 2.2 & 8.9 ± 2.1  & 35.7 ± 1.3 & 72.9 ± 6.3 & 20.1 ± 3.8  \\ 
    \textbf{ECP($\lambda$=1.0)}* & 84.4 ± 2.3 & 9.0 ± 2.0  & 35.9 ± 1.3 & 73.8 ± 5.4 & 21.0 ± 3.7  \\ 
    ECP($\lambda$=10.0)          & 84.3 ± 2.7 & 9.2 ± 2.4  & 35.8 ± 1.4 & 73.5 ± 6.0 & 20.6 ± 4.3  \\
    ECP($\lambda$=100.0)         & 70.8 ± 3.9 & 18.6 ± 2.8 & 21.4 ± 2.7 & 58.7 ± 2.6 & 28.9 ± 7.1  \\
    \cdashline{\CA-\setCOL}
    LS($\alpha$=0.01)            & 83.4 ± 2.8 & 9.0 ± 2.9 & 32.9 ± 0.1 & 66.1 ± 5.7 & 18.4 ± 3.6  \\
    \textbf{LS($\alpha$=0.05)}*  & 83.0 ± 3.0 & 7.5 ± 2.2 & 25.1 ± 0.5 & 62.6 ± 3.3 & 17.5 ± 4.0   \\
    LS($\alpha$=0.1)             & 84.1 ± 2.3 & 8.4 ± 2.9 & 17.5 ± 0.1 & 62.3 ± 2.5 & 18.5 ± 3.5   \\ 
    \cdashline{\CA-\setCOL}
    SVLS($\sigma$=1)*           & 83.9 ± 2.5 & 9.0 ± 2.3 & 32.6 ± 1.1 & 69.6 ± 8.3 & 18.8 ± 2.8  \\ 
    \textbf{SVLS($\sigma$=2)}   & 84.2 ± 2.6 & 9.0 ± 2.0 & 32.2 ± 1.1 & 70.8 ± 7.1 & 18.1 ± 3.5   \\ 
    SVLS($\sigma$=3)            & 83.9 ± 2.7 & 9.0 ± 2.2 & 32.1 ± 1.2 & 69.3 ± 7.0 & 18.8 ± 2.8  \\
    \cdashline{\CA-\setCOL}
    \textbf{MbLS($\lambda=0.1$,m=30)} & 83.7 ± 2.6 & 9.0 ± 2.0  & 35.4 ± 1.2 & 70.0 ± 5.6 & 19.7 ± 4.1  \\
    MbLS($\lambda=0.1$,m=20)*         & 84.0 ± 2.6 & 9.2 ± 2.1  & 35.3 ± 1.3 & 67.5 ± 5.7 & 19.5 ± 3.5  \\
    MbLS($\lambda=0.1$,m=10)          & 82.4 ± 2.6 & 9.8 ± 2.8  & 33.1 ± 1.2 & 64.1 ± 7.0 & 18.3 ± 3.1  \\
    MbLS($\lambda=0.1$,m=8)           & 62.4 ± 8.2 & 18.9 ± 1.6 & 26.3 ± 0.4 & 73.6 ± 6.6 & 38.3 ± 4.1  \\
    \hdashline[0.5pt/5pt]
    \textbf{MbLS($\lambda=1$,m=20)}  & 83.4 ± 2.5 & 9.2 ± 2.6 & 35.4 ± 1.3 & 71.3 ± 7.0 & 20.1 ± 3.6  \\
    MbLS($\lambda=10$,m=20)          & 83.0 ± 3.4 & 9.5 ± 2.8 & 34.6 ± 1.4 & 69.1 ± 6.0 & 20.1 ± 4.1   \\ 
    MbLS($\lambda=100$,m=20)         & 82.5 ± 3.2 & 9.1 ± 3.0 & 32.4 ± 1.4 & 68.2 ± 7.3 & 19.0 ± 2.9 \\
    \cdashline{\CA-\setCOL}
    TTA                      & 84.1 ± 2.8 & 9.1 ± 2.1 & 35.5 ± 1.4 & 72.9 ± 5.9 & 20.8 ± 3.9  \\ 
    \cdashline{\CA-\setCOL}
    Ens                    & 85.0 ± 2.7 & 7.8 ± 1.9 & 34.5 ± 1.2 & 78.6 ± 4.7 & 25.7 ± 6.8   \\ 
    \cdashline{\CA-\setCOL}
    Bayes                     & 83.9 ± 2.6 & 8.7 ± 2.1 & 34.5 ± 1.2 & 74.1 ± 5.4 & 22.1 ± 3.5  \\
    Bayes + 10AvU             & 83.4 ± 2.8 & 8.7 ± 2.4 & 34.7 ± 1.3 & 74.7 ± 4.9 & 24.4 ± 4.1   \\
    \textbf{Bayes + 100AvU}*  & 83.6 ± 2.5 & 7.6 ± 2.5 & 35.6 ± 1.2 & 76.1 ± 5.6 & 25.1 ± 5.3   \\
    Bayes + 1000AvU           & 83.5 ± 3.0 & 8.5 ± 3.4 & 36.1 ± 1.5 & 77.2 ± 6.0 & 24.7 ± 4.5   \\
     
     \hline
\end{tabular}
\end{table}

\begin{table}[H]
\centering
\caption{Volumetric (DICE), calibrative (ECE) and uncertainty-error correspondence metrics (AvU-AUC, ROC-AUC, PRC-AUC) on head-and-neck OOD dataset. The experiment indicated as \textbf{bold} is the one with the best performance. * indicates hyperparameters chosen by the validation dataset.}
\begin{tabular}{|l|c|c|c|c|c|}
    \hline
    \thead{\textbf{Experiment}} & \thead{DICE $\uparrow$ \\(x$10^{-2}$)}  & \thead{ECE $\downarrow$ \\(x$10^{-2}$)}  & \thead{AvU-AUC $\uparrow$ \\(x$10^{-2}$)} & \thead{ROC-AUC $\uparrow$\\(x$10^{-2}$)} & \thead{PRC-AUC $\uparrow$\\(x$10^{-2}$)} \\ 
    \hline
    Det                     & 78.1 ± 4.6 & 12.9 ± 2.6 & 33.4 ± 1.4 & 62.2 ± 4.5 & 24.1 ± 3.7  \\
    Det + 10AvU             & 76.3 ± 6.9 & 13.7 ± 3.5 & 33.3 ± 1.7 & 58.3 ± 4.6 & 23.3 ± 4.4   \\
    \textbf{Det + 100AvU}*  & 78.6 ± 4.7 & 12.7 ± 3.0 & 34.2 ± 1.5 & 60.8 ± 4.7 & 22.4 ± 4.1    \\
    Det + 1000AvU           & 42.5 ± 7.2 & 12.1 ± 2.1 & 28.9 ± 1.7 & 66.1 ± 5.8 & 19.0 ± 6.2  \\
    \cdashline{\CA-\setCOL}
    \textbf{Focal($\gamma$=1)}* & 77.2 ± 6.7 & 12.5 ± 2.9 & 30.6 ± 1.7 & 57.0 ± 4.6 & 20.9 ± 4.2   \\ 
    Focal($\gamma$=2)           & 77.7 ± 5.2 & 12.2 ± 1.9 & 24.1 ± 0.9 & 57.5 ± 4.6 & 21.0 ± 4.1   \\ 
    Focal($\gamma$=3)           & 79.0 ± 5.2 & 13.3 ± 1.6 & 18.6 ± 0.7 & 59.8 ± 4.9 & 16.6 ± 3.9  \\ 
    \cdashline{\CA-\setCOL}
    ECP($\lambda$=0.1)           & 78.5 ± 4.9 & 12.6 ± 2.8 & 33.5 ± 1.6  & 59.8 ± 4.9 & 22.0 ± 3.8  \\ 
    \textbf{ECP($\lambda$=1.0)}* & 78.8 ± 4.3 & 12.5 ± 2.6 & 36.6 ± 1.5  & 61.5 ± 4.8 & 23.2 ± 3.6  \\ 
    ECP($\lambda$=10.0)          & 78.9 ± 4.5 & 12.4 ± 2.5 & 33.8 ± 1.5  & 60.1 ± 4.7 & 22.1 ± 3.5  \\
    ECP($\lambda$=100.0)         & 62.0 ± 6.1 & 20.0 ± 1.8 & 19.9 ± 2.9  & 56.0 ± 2.8 & 36.5 ± 9.7  \\
    \cdashline{\CA-\setCOL}
    LS($\alpha$=0.1)            & 77.7 ± 6.0 & 8.9 ± 2.7  & 17.9 ± 0.3  & 57.6 ± 1.9 & 23.9 ± 4.4   \\ 
    \textbf{LS($\alpha$=0.05)}* & 77.7 ± 6.0 & 10.3 ± 2.9 & 24.3 ± 0.7  & 56.7 ± 3.3 & 20.6 ± 4.3   \\ 
    LS($\alpha$=0.01)           & 77.9 ± 5.4 & 13.3 ± 2.8 & 31.1 ± 1.5  & 58.6 ± 3.9 & 22.4 ± 3.7  \\
    \cdashline{\CA-\setCOL}
    SVLS($\sigma$=1)*           & 78.3 ± 6.1 & 11.5 ± 3.0 & 31.4 ± 1.4 & 61.1 ± 4.9 & 23.3 ± 3.3  \\ 
    \textbf{SVLS($\sigma$=2)}   & 79.0 ± 6.0 & 11.3 ± 2.5 & 31.4 ± 1.2 & 59.9 ± 5.4 & 21.6 ± 2.7   \\ 
    SVLS($\sigma$=3)            & 78.6 ± 5.1 & 11.5 ± 2.9 & 31.1 ± 1.5 & 58.7 ± 5.0 & 22.5 ± 3.8  \\ 
    \cdashline{\CA-\setCOL}
    MbLS($\lambda=0.1$,m=30)           & 76.5 ± 7.1  & 13.6 ± 3.9 & 32.1 ± 2.9 & 58.9 ± 4.1 & 24.7 ± 7.7  \\
    \textbf{MbLS($\lambda=0.1$,m=20)}* & 77.5 ± 6.3  & 13.4 ± 3.0 & 33.4 ± 1.5 & 56.9 ± 5.0 & 21.5 ± 3.6  \\
    MbLS($\lambda=0.1$,m=10)           & 76.8 ± 6.3  & 13.0 ± 3.2 & 31.7 ± 1.4 & 53.0 ± 4.5 & 20.6 ± 3.9  \\
    MbLS($\lambda=0.1$,m=8)            & 50.3 ± 10.6 & 20.1 ± 2.8 & 26.2 ± 0.9 & 61.1 ± 7.1 & 34.1 ± 3.7  \\ 
    \hdashline[0.5pt/5pt]
    MbLS($\lambda=1$,m=20)             & 77.3 ± 6.2  & 13.2 ± 2.8 & 33.3 ± 1.6 & 61.0 ± 4.5 & 23.4 ± 4.1  \\
    \textbf{MbLS($\lambda=10$,m=20)}   & 78.1 ± 5.3  & 13.0 ± 2.9 & 32.9 ± 1.5 & 57.0 ± 4.1 & 21.7 ± 3.5  \\
    MbLS($\lambda=100$,m=20)           & 78.2 ± 4.9  & 12.7 ± 2.5 & 31.6 ± 1.3 & 55.0 ± 5.1 & 19.7 ± 3.5 \\
    \cdashline{\CA-\setCOL}
    TTA                      & 78.1 ± 4.6 & 12.7 ± 2.6 & 33.2 ± 1.5  & 62.7 ± 4.6 & 24.9 ± 4.1  \\ 
    \cdashline{\CA-\setCOL}
    Ens                    & 78.6 ± 5.2 & 10.6 ± 2.4 & 32.1 ± 1.9 & 64.7 ± 4.9 & 28.2 ± 5.1   \\ 
    \cdashline{\CA-\setCOL}
    Bayes                     & 75.0 ± 9.9 & 12.4 ± 4.0 & 32.2 ± 1.8 & 64.8 ± 5.0 & 27.7 ± 5.8  \\
    Bayes + 10AvU             & 74.9 ± 9.5 & 12.4 ± 4.0 & 32.1 ± 2.0 & 65.2 ± 4.6 & 29.1 ± 6.1   \\
    \textbf{Bayes + 100AvU}*  & 76.3 ± 7.7 & 12.1 ± 3.7 & 33.2 ± 1.7 & 65.8 ± 5.0 & 30.1 ± 6.5   \\
    Bayes + 1000AvU           & 75.5 ± 8.2 & 14.3 ± 4.1 & 33.5 ± 1.8 & 69.3 ± 5.6 & 32.9 ± 6.9   \\
     
     \hline
\end{tabular}
\end{table}

%%%%%%%%%%%%%%%%%%%%%%%%%%%%%%%%%%%%
% Prostate
%%%%%%%%%%%%%%%%%%%%%%%%%%%%%%%%%%%%

\begin{table}[H]
\centering
\caption{Volumetric (DICE), calibrative (ECE) and uncertainty-error correspondence metrics (AvU-AUC, ROC-AUC, PRC-AUC) on prostate validation dataset for the purpose of hyperparameter selection. The experiment indicated as \textbf{bold} is the one with the best performance.}
\begin{tabular}{|l|c|c|c|c|c|}
    \hline
    \thead{\textbf{Experiment}} & \thead{DICE $\uparrow$ \\(x$10^{-2}$)}  & \thead{ECE $\downarrow$ \\(x$10^{-2}$)}  & \thead{AvU-AUC $\uparrow$ \\(x$10^{-2}$)} & \thead{ROC-AUC $\uparrow$\\(x$10^{-2}$)} & \thead{PRC-AUC $\uparrow$\\(x$10^{-2}$)} \\ 
    \hline
    Det                     & 85.9 ± 1.8 & 14.4 ± 3.2 & 96.5 ± 0.9 & 92.6 ± 4.1 & 26.5 ± 1.5 \\
    Det + 100AvU            & 84.8 ± 2.3 & 16.3 ± 3.9 & 96.1 ± 0.9 & 91.7 ± 4.2 & 27.9 ± 2.8  \\
    \textbf{Det + 1000AvU}  & 84.8 ± 1.9 & 16.0 ± 3.0 & 96.4 ± 0.9 & 93.6 ± 3.2 & 29.2 ± 1.0  \\
    Det + 10000AvU          & 84.9 ± 3.5 & 16.7 ± 5.1 & 96.5 ± 1.0 & 91.8 ± 2.7 & 25.9 ± 2.3  \\
    \cdashline{\CA-\setCOL}
    Ensemble                & 85.4 ± 1.7 & 13.4 ± 3.0 & 96.0 ± 1.0 & 94.8 ± 2.4 & 31.4 ± 1.6  \\ 
    \cdashline{\CA-\setCOL}
    \textbf{Focal($\gamma$=1)} & 84.5 ± 2.7 & 13.3 ± 4.3 & 90.7 ± 1.1 & 93.0 ± 4.1 & 29.4 ± 1.7\\ 
    Focal($\gamma$=2)          & 84.4 ± 2.1 & 9.8 ± 2.6  & 82.5 ± 1.0 & 93.8 ± 2.3 & 30.9 ± 2.1 \\ 
    Focal($\gamma$=3)          & 84.5 ± 1.9 & 6.4 ± 1.5  & 58.9 ± 1.3 & 92.0 ± 4.3 & 30.5 ± 2.6 \\
    % Focal($\gamma$=4)   & 84.3 ± 1.8 & 6.6 ± 1.5  & 60.4 ± 1.3 & 94.4 ± 2.7 & 31.4 ± 3.0 \\
    \cdashline{\CA-\setCOL}
    ECP($\lambda$=0.1)            & 85.9 ± 1.8 & 14.6 ± 3.0 & 96.5 ± 0.9 & 91.9 ± 4.2 & 25.8 ± 1.7  \\ 
    ECP($\lambda$=1.0)            & 85.7 ± 1.8 & 14.7 ± 3.0 & 96.4 ± 1.0 & 92.3 ± 3.9 & 26.4 ± 1.7  \\ 
    \textbf{ECP($\lambda$=10.0)}  & 85.7 ± 1.7 & 14.8 ± 2.7 & 96.4 ± 1.0 & 91.9 ± 4.5 & 26.0 ± 1.8  \\ 
    ECP($\lambda$=100.0)          & 85.7 ± 1.8 & 14.8 ± 2.8 & 96.4 ± 1.0 & 91.8 ± 4.3 & 25.8 ± 1.9  \\ 
    ECP($\lambda$=1000.0)         & 86.0 ± 1.9 & 15.0 ± 3.0 & 85.0 ± 0.3 & 88.7 ± 2.1 & 26.7 ± 3.4  \\ 
    \cdashline{\CA-\setCOL}
    LS($\alpha$=0.01)          & 83.7 ± 2.5 & 17.2 ± 4.1 & 91.9 ± 0.9 & 85.8 ± 5.4 & 28.2 ± 2.9  \\
    \textbf{LS($\alpha$=0.05)} & 85.1 ± 1.4 & 13.6 ± 2.3 & 80.8 ± 0.9 & 84.3 ± 5.7 & 25.4 ± 2.2  \\ 
    LS($\alpha$=0.1)           & 85.0 ± 2.1 & 11.1 ± 3.4 & 70.3 ± 0.6 & 85.1 ± 3.6 & 27.0 ± 2.2  \\ 
    \cdashline{\CA-\setCOL}
    SVLS($\sigma$=1)           & 84.5 ± 1.9 & 14.0 ± 2.6 & 92.4 ± 1.0 & 91.8 ± 2.3 & 22.9 ± 1.8  \\ 
    \textbf{SVLS($\sigma$=2)}  & 85.0 ± 1.8 & 12.9 ± 3.1 & 92.4 ± 0.9 & 91.4 ± 3.0 & 22.1 ± 1.4  \\ 
    SVLS($\sigma$=3)           & 85.0 ± 1.6 & 13.1 ± 2.7 & 92.1 ± 0.9 & 91.2 ± 2.5 & 21.9 ± 1.4  \\ 
    \cdashline{\CA-\setCOL}
    \textbf{MbLS($\lambda=0.1$,m=10)}  & 84.8 ± 1.4 & 17.5 ± 5.1 & 95.7 ± 1.1 & 91.2 ± 4.1 & 31.1 ± 1.7  \\
    MbLS($\lambda=0.1$,m=8)            & 83.8 ± 1.3 & 16.0 ± 2.2 & 93.9 ± 0.9 & 90.5 ± 3.5 & 27.9 ± 2.1  \\ 
    MbLS($\lambda=0.1$,m=5)            & 84.3 ± 1.6 & 15.5 ± 2.8 & 90.4 ± 0.8 & 90.1 ± 5.6 & 28.2 ± 2.2 \\ 
    MbLS($\lambda=0.1$,m=3)            & 84.2 ± 2.1 & 12.8 ± 3.3 & 70.8 ± 0.4 & 82.1 ± 3.5 & 28.8 ± 5.6  \\ 
    % MbLS(m=20)   & 84.8 ± 3.0 & 16.3 ± 4.9 & 96.2 ± 1.0 & ± & ±    \\
    \hdashline[0.5pt/5pt]
    \textbf{MbLS($\lambda=1.0$,m=10)}  & 83.7 ± 1.2 & 17.4 ± 4.4 & 96.0 ± 1.0 & 91.2 ± 3.9 & 30.5 ± 1.5  \\
    MbLS($\lambda=10.0$,m=10)          & 83.9 ± 1.5 & 17.8 ± 4.1 & 95.0 ± 1.2 & 90.8 ± 3.6 & 30.9 ± 1.6 \\
    \cdashline{\CA-\setCOL}
    TTA                   & 85.6 ± 1.7 & 14.5 ± 3.1 & 96.3 ± 0.9 & 92.5 ± 4.0 & 27.2 ± 1.6 \\ 
    \cdashline{\CA-\setCOL}
    Bayes                    & 85.7 ± 2.3 & 10.7 ± 3.0 & 93.2 ± 1.8 & 95.3 ± 1.9 & 30.3 ± 2.9  \\
    Bayes + 100AvU           & 86.1 ± 3.0 & 11.5 ± 3.8 & 94.9 ± 2.1 & 95.9 ± 2.0 & 31.5 ± 3.5   \\
    \textbf{Bayes + 1000AvU} & 85.8 ± 2.8 & 12.0 ± 3.9 & 95.5 ± 1.9 & 96.3 ± 2.4 & 32.0 ± 3.3  \\
    Bayes + 10000AvU         & 86.0 ± 2.4 & 10.9 ± 3.0 & 96.1 ± 1.7 & 93.1 ± 2.1 & 29.3 ± 3.1 \\
     
     \hline
\end{tabular}
\end{table}

\begin{table}[H]
\centering
\caption{Volumetric (DICE), calibrative (ECE) and uncertainty-error correspondence metrics (AvU-AUC, ROC-AUC, PRC-AUC) on prostate ID dataset. The experiment indicated as \textbf{bold} is the one with the best performance.  * indicates hyperparameters chosen by the validation dataset.}
\begin{tabular}{|l|c|c|c|c|c|}
    \hline
    \thead{\textbf{Experiment}} & \thead{DICE $\uparrow$ \\(x$10^{-2}$)}  & \thead{ECE $\downarrow$ \\(x$10^{-2}$)}  & \thead{AvU-AUC $\uparrow$ \\(x$10^{-2}$)} & \thead{ROC-AUC $\uparrow$\\(x$10^{-2}$)} & \thead{PRC-AUC $\uparrow$\\(x$10^{-2}$)} \\ 
    \hline
    Det                     & 84.1 ± 5.6 & 12.9 ± 6.0 & 96.1 ± 3.4 & 92.5 ± 5.7 & 28.0 ± 3.7  \\
    Det + 100AvU            & 83.7 ± 6.7 & 16.6 ± 7.2 & 95.7 ± 3.3 & 91.6 ± 6.2 & 27.2 ± 2.9   \\
    \textbf{Det + 1000AvU}* & 83.7 ± 6.8 & 16.9 ± 8.1 & 95.9 ± 3.8 & 92.1 ± 6.8 & 28.2 ± 3.4    \\
    Det + 10000AvU          & 83.4 ± 6.4 & 18.1 ± 7.9 & 96.1 ± 3.7 & 90.7 ± 5.6 & 26.1 ± 3.4  \\
    \cdashline{\CA-\setCOL}
    \textbf{Focal ($\gamma$=1)}* & 81.1 ± 15.4 & 10.2 ± 5.0 & 90.3 ± 0.3 &  93.2 ± 5.5 & 29.3 ± 3.4   \\ 
    Focal ($\gamma$=2)           & 83.1 ± 6.2 & 10.4 ± 6.8  & 81.6 ± 2.5 &  92.9 ± 5.3 & 30.1 ± 3.7   \\ 
    Focal ($\gamma$=3)           & 82.3 ± 7.2 & 8.0 ± 6.4   & 58.7 ± 1.2 &  92.5 ± 5.4 & 31.8 ± 3.5  \\ 
    \cdashline{\CA-\setCOL}
    ECP ($\lambda$=0.1)             & 84.1 ± 5.4 & 16.5 ± 7.0  & 96.1 ± 3.4 &  92.3 ± 6.0 & 27.6 ± 3.9  \\ 
    ECP ($\lambda$=1.0)             & 84.1 ± 5.5 & 16.4 ± 7.0  & 96.1 ± 3.3 &  92.3 ± 6.0 & 27.8 ± 4.3  \\ 
    \textbf{ECP ($\lambda$=10.0)}*  & 84.0 ± 5.5 & 16.7 ± 7.1  & 96.1 ± 3.4 &  92.1 ± 6.0 & 27.6 ± 4.3  \\ 
    ECP ($\lambda$=100.0)           & 84.0 ± 5.5 & 16.6 ± 7.0  & 96.0 ± 3.0 &  92.1 ± 6.0 & 27.6 ± 4.1  \\ 
    ECP ($\lambda$=1000.0)          & 84.1 ± 5.7 & 16.6 ± 7.0  & 86.1 ± 3.2 &  92.2 ± 5.9 & 27.5 ± 4.3  \\ 
    \cdashline{\CA-\setCOL}
    LS ($\alpha$=0.01)           & 82.5 ± 8.3 & 18.0 ± 9.4 & 91.3 ± 3.9 & 86.2 ± 7.9 & 27.0 ± 3.8 \\
    \textbf{LS ($\alpha$=0.05)}* & 83.4 ± 7.2 & 15.1 ± 8.6 & 80.4 ± 2.9 & 83.2 ± 7.8 & 25.1 ± 3.1 \\ 
    LS ($\alpha$=0.1)            & 84.1 ± 5.6 & 11.6 ± 7.0 & 70.1 ± 1.8 & 84.7 ± 6.2 & 26.9 ± 3.3 \\ 
    \cdashline{\CA-\setCOL}
    SVLS ($\sigma$=1)           & 83.4 ± 7.1 & 14.7 ± 8.8 & 92.0 ± 3.7 &  90.9 ± 7.4 & 22.9 ± 2.9  \\ 
    \textbf{SVLS ($\sigma$=2)}* & 83.5 ± 6.7 & 14.0 ± 8.1 & 91.9 ± 4.1 &  90.5 ± 7.9 & 21.7 ± 2.6   \\ 
    SVLS($\sigma$=3)            & 83.2 ± 8.1 & 14.3 ± 9.7 & 91.5 ± 3.9 &  91.0 ± 6.8 & 23.1 ± 3.1  \\ 
    \cdashline{\CA-\setCOL}
    MbLS ($\lambda=1.0$,m=3)           & 83.2 ± 6.3 & 13.3 ± 7.8 & 70.6 ± 1.7 &  82.2 ± 6.3 & 27.7 ± 3.4  \\  
    MbLS ($\lambda=1.0$,m=5)           & 82.8 ± 6.6 & 16.7 ± 8.0 & 89.9 ± 3.2 &  90.5 ± 7.2 & 27.2 ± 4.4  \\
    \textbf{MbLS ($\lambda=1.0$,m=8)}  & 83.5 ± 5.8 & 17.1 ± 7.0 & 95.3 ± 3.6 &  93.0 ± 5.2 & 27.8 ± 4.1  \\
    MbLS ($\lambda=1.0$,m=10)          & 84.2 ± 5.3 & 18.1 ± 6.1 & 95.5 ± 3.3 &  91.7 ± 6.1 & 26.5 ± 3.5  \\ 
    % MbLS (m=20)   & 83.0 ± 7.4 & 17.8 ± 8.8 & 95.6 ± 3.9 &  90.3 ± 6.6 & 26.2 ± 4.6  \\
    \hdashline[0.5pt/5pt]
    \textbf{MbLS($\lambda=1.0$,m=10)}* & 84.2 ± 4.9 & 17.9 ± 7.4 & 95.6 ± 2.9 &  92.2 ± 5.6 & 26.9 ± 3.6  \\
    MbLS($\lambda=10.0$,m=10)          & 83.9 ± 5.2 & 17.9 ± 8.0 & 95.1 ± 3.2 &  91.9 ± 5.9 & 26.2 ± 4.1 \\
    \cdashline{\CA-\setCOL}
    TTA                    & 83.8 ± 5.8 & 16.4 ± 7.1 & 96.0 ± 3.5 & 92.7 ± 5.6 & 28.8 ± 3.9  \\ 
    \cdashline{\CA-\setCOL}
    Ensemble               & 84.5 ± 5.7 & 11.3 ± 6.5 & 95.2 ± 3.5 &  94.3 ± 4.3 & 30.0 ± 4.6   \\ 
    \cdashline{\CA-\setCOL}
    Bayes                      & 84.0 ± 5.8 & 8.6 ± 4.7  & 92.1 ± 2.6 &  94.7 ± 3.1 & 29.1 ± 4.8  \\
    Bayes + 100AvU             & 84.1 ± 6.4 & 12.0 ± 6.2 & 94.4 ± 3.1 &  95.5 ± 2.9 & 28.9 ± 5.0   \\
    \textbf{Bayes + 1000AvU}*  & 84.9 ± 6.9 & 8.9 ± 6.0  & 94.5 ± 3.2 &  95.7 ± 3.2 & 30.5 ± 4.5   \\
    Bayes + 10000AvU           & 85.2 ± 5.9 & 11.0 ± 6.3 & 94.2 ± 3.6 &  95.9 ± 3.5 & 30.2 ± 4.0   \\ 
    \hline
\end{tabular}
\end{table}

\begin{table}[H]
\centering
\caption{Volumetric (DICE), calibrative (ECE) and uncertainty-error correspondence metrics (AvU-AUC, ROC-AUC, PRC-AUC) on prostate OOD dataset. The experiment indicated as \textbf{bold} is the one with the best performance. * indicates hyperparameters chosen by the validation dataset.}
\begin{tabular}{|l|c|c|c|c|c|}
    \hline
    \thead{\textbf{Experiment}} & \thead{DICE $\uparrow$ \\(x$10^{-2}$)}  & \thead{ECE $\downarrow$ \\(x$10^{-2}$)}  & \thead{AvU-AUC $\uparrow$ \\(x$10^{-2}$)} & \thead{ROC-AUC $\uparrow$\\(x$10^{-2}$)} & \thead{PRC-AUC $\uparrow$\\(x$10^{-2}$)} \\ 
    \hline
    Det                      & 74.2 ± 12.6 & 15.6 ± 6.3  & 92.3 ± 5.4 & 87.9 ± 7.5 & 22.1 ± 6.2  \\
    Det + 100AvU             & 74.2 ± 13.3 & 23.6 ± 11.2 & 93.0 ± 4.2 &  87.1 ± 6.2 & 22.2 ± 5.7   \\
    \textbf{Det + 1000AvU}*  & 74.5 ± 13.0 & 27.6 ± 14.3 & 92.2 ± 5.7 &  88.2 ± 7.6 & 22.0 ± 7.1    \\
    Det + 10000AvU           & 72.7 ± 15.1 & 27.6 ± 14.3 & 92.4 ± 5.2 &  82.3 ± 9.4 & 19.6 ± 6.2  \\
    \cdashline{\CA-\setCOL}
    \textbf{Focal($\gamma$=1)}* & 71.2 ± 17.4 & 12.1 ± 5.8 & 85.4 ± 6.1 &  89.0 ± 7.1 & 24.3 ± 6.7   \\ 
    Focal($\gamma$=2)           & 76.7 ± 10.8 & 12.8 ± 8.2 & 72.0 ± 9.3 &  87.2 ± 7.6 & 22.4 ± 6.4   \\ 
    Focal($\gamma$=3)           & 73.2 ± 13.7 & 11.6 ± 7.7 & 49.7 ± 9.4 &  87.1 ± 8.5 & 27.0 ± 7.2  \\ 
    \cdashline{\CA-\setCOL}
    ECP($\lambda$=0.1)             & 74.6 ± 12.5 & 22.8 ± 10.5  & 92.1 ± 5.5 &  87.6 ± 7.6 & 21.3 ± 6.6  \\ 
    ECP($\lambda$=1.0)             & 73.9 ± 13.1 & 23.2 ± 10.7  & 91.9 ± 5.6 &  87.2 ± 7.2 & 21.2 ± 6.4  \\ 
    \textbf{ECP($\lambda$=10.0)}*  & 74.8 ± 12.5 & 22.3 ± 10.2  & 91.6 ± 6.3 &  87.2 ± 8.1 & 20.6 ± 7.0  \\ 
    ECP($\lambda$=100.0)           & 74.9 ± 12.3 & 22.7 ± 10.5  & 92.1 ± 5.5 &  87.7 ± 8.0 & 21.5 ± 7.2  \\ 
    ECP($\lambda$=1000.0)          & 74.6 ± 12.5 & 22.7 ± 10.3  & 92.2 ± 5.6 &  87.6 ± 7.7 & 21.5 ± 6.7  \\
    \cdashline{\CA-\setCOL}
    LS($\alpha$=0.01)           & 71.6 ± 15.1 & 24.6 ± 11.6 & 87.9 ± 5.3 & 84.3 ± 7.5 & 22.7 ± 6.2  \\
    \textbf{LS($\alpha$=0.05)}* & 74.5 ± 13.0 & 21.7 ± 11.5 & 77.2 ± 4.6 & 79.5 ± 8.9 & 19.1 ± 7.2  \\ 
    LS($\alpha$=0.1)            & 75.2 ± 12.2 & 18.1 ± 10.1 & 67.4 ± 3.8 & 79.0 ± 8.4 & 19.9 ± 6.4  \\ 
    \cdashline{\CA-\setCOL}
    SVLS($\sigma$=1)           & 74.9 ± 11.7 & 19.7 ± 9.1  & 88.5 ± 5.5 &  87.2 ± 7.4 & 18.7 ± 5.1  \\ 
    \textbf{SVLS($\sigma$=2)}* & 76.9 ± 11.5 & 17.9 ± 9.3  & 88.3 ± 5.2 &  87.2 ± 7.2 & 16.4 ± 5.2  \\ 
    SVLS($\sigma$=3)           & 74.3 ± 13.5 & 21.4 ± 12.6 & 88.4 ± 5.1 &  86.3 ± 8.2 & 19.4 ± 5.0  \\ 
    \cdashline{\CA-\setCOL}
    % MbLS(m=20)          & 75.2 ± 13.3 & 22.0 ± 11.2 & ± &  85.4 ± 8.4  & 19.3 ± 8.1  \\
    \textbf{MbLS($\lambda=0.1$, m=10)}  & 72.3 ± 15.9 & 20.9 ± 7.9  & 91.4 ± 5.7 & 87.9 ± 6.9  & 22.2 ± 6.7  \\ 
    MbLS($\lambda=0.1$, m=8)            & 74.1 ± 13.5 & 20.7 ± 8.7  & 88.3 ± 8.2 & 85.0 ± 10.4 & 18.8 ± 8.8  \\ 
    MbLS($\lambda=0.1$, m=5)            & 74.7 ± 13.3 & 22.0 ± 11.3 & 86.9 ± 5.0 & 87.1 ± 7.9  & 22.0 ± 6.4  \\ 
    MbLS($\lambda=0.1$, m=3)            & 74.0 ± 13.3 & 20.5 ± 11.7 & 68.6 ± 2.9 & 78.0 ± 7.2  & 21.5 ± 6.7  \\
    \hdashline[0.5pt/5pt]
    \textbf{MbLS($\lambda=1.0$, m=10)}* & 73.6 ± 12.5 & 19.9 ± 7.4 & 91.8 ± 3.4 & 86.5 ± 7.2 & 21.8 ± 5.6\\
    MbLS($\lambda=10.0$, m=10)          & 72.1 ± 16.1 & 20.2 ± 6.7 & 91.4 ± 5.5 & 86.5 ± 9.0 & 22.2 ± 6.7\\
    \cdashline{\CA-\setCOL}
    TTA                       & 74.0 ± 12.8 & 23.7 ± 11.4 & 92.8 ± 4.8 & 88.6 ± 7.4 & 24.9 ± 5.8  \\ 
    \cdashline{\CA-\setCOL}
    Ensemble                  & 76.3 ± 12.2 & 9.7 ± 5.0 & 89.9 ± 6.6 & 91.6 ± 5.2 & 28.4 ± 5.7   \\ 
    \cdashline{\CA-\setCOL}
    Bayes                     & 70.6 ± 16.6 & 11.8 ± 7.2  & 86.2 ± 6.0 &  89.1 ± 7.4 & 25.7 ± 5.1  \\
    Bayes + 100AvU            & 72.1 ± 14.4 & 20.0 ± 11.8 & 91.0 ± 3.8 &  92.7 ± 4.0 & 30.2 ± 6.5  \\
    \textbf{Bayes + 1000AvU}* & 76.3 ± 12.6 & 11.4 ± 6.7  & 89.5 ± 6.2 &  90.6 ± 6.9 & 26.2 ± 7.4  \\
    Bayes + 10000AvU          & 76.6 ± 12.7 & 17.1 ± 10.1 & 88.6 ± 6.5 &  90.4 ± 6.3 & 23.3 ± 7.4  \\ 
     \hline
\end{tabular}
\end{table}

\newpage
\section{Visual Results}
\label{sec:app_visualresults}
Visual results in \blueautoref{fig:results_visual_hn} and \blueautoref{fig:results_visual_pros} show pairs of consecutive CT/MR slices to better understand the 3D nature of the output uncertainty across all models. We show examples with both high and low DICE to investigate the presence and absence of uncertainty in different regions of the model prediction. 

\subsection{Head-And-Neck CT}
The first two rows of \blueautoref{fig:results_visual_hn_rtog} and \blueautoref{fig:results_visual_hn_structseg} show the mandible (i.e. lower jaw bone) with only the \textit{Bayes+AvU} model having overall low uncertainty in accurate regions and high uncertainty in (or close to) inaccurate regions. 

In the next set of rows for head-and-necks CTs, we observe the parotid gland, a salivary organ, with (\blueautoref{fig:results_visual_hn_rtog} - Case 2) and without (\blueautoref{fig:results_visual_hn_structseg} - Case 2, Case 3) a dental scattering issue. In both cases, while the \textit{Det} model shows low uncertainty, the baseline \textit{Bayes} model shows high uncertainty in accurate regions. Usage of the AvU loss lowers uncertainty in these regions, while still exhibiting uncertainty in the erroneous regions, for e.g. the medial (i.e. internal) portion of the organ in \blueautoref{fig:results_visual_hn_rtog} (Case 2). 

Moving on to our last case, we see the submandibular gland, another salivary gland in \blueautoref{fig:results_visual_hn_rtog} (Case 3). The \textit{Ensemble}, \textit{Focal}, \textit{SVLS} and \textit{MBLS} models all display high uncertainty in the core of the organ, which are also accurately predicted. On the other hand, the AvU loss minimizes the uncertainty and shows uncertainty in the erroneous region on the second slice.

\subsection{Prostate MR}
For the prostate datasets, we see two cases with high DICE in \blueautoref{fig:results_visual_pros_prmeddec} (Case 1) and \blueautoref{fig:results_visual_pros_pr12} (Case 2) where the use of the AvU loss reduces uncertainty for the baseline \textit{Bayes} model. 

We also see cases with low DICE in \blueautoref{fig:results_visual_pros_prmeddec} (Case 2) and \blueautoref{fig:results_visual_pros_pr12} (Case 1). Due to their low DICE all models display high uncertainty, but the \textit{Bayes+AvU} model shows high overlap between its uncertain and erroneous regions. The same is also observed in \blueautoref{fig:results_visual_pros_pr12} (Case 3). 

Finally, in \blueautoref{fig:results_visual_pros_prmeddec} (Case 3), we do not see any clear benefit of using the AvU loss on the \textit{Bayes} model.

\newpage 
\section{BayesH model}
\label{sec:app_bayesh}

% Table-VSM
\begin{table}[!tbh]
    \centering
    \caption{Volumetric (\textit{DICE}) , calibrative (\textit{ECE}) and uncertainty-error correspondence metrics (ROC-AUC, PRC-AUC) for different Bayesian models. We evaluate head-and-neck (H\&N) CT and Prostate MR test datasets which are either in-distribution (ID) or out-of-distribution (OOD). The arrows in the table header indicate whether a metric should be high ($\uparrow$) or low ($\downarrow$). Here, $^{\dag}$ and \textbf{bold} are used to indicate a statistical significance and improved results upon comparing a Bayesian model and its AvU-loss version, while \underline{underlined} numbers indicate the best value for a metric across a dataset.}
    
    \begin{tabular}{|c|c|c|c|c|c|}
    
    \hline
    \thead{Test \\ Dataset} & \thead{Model} & \thead{DICE $\uparrow$\\(x$10^{-2}$)} & \thead{ECE $\downarrow$\\(x$10^{-2}$)} & \thead{ROC-AUC $\uparrow$\\(x$10^{-2}$)} & \thead{PRC-AUC$\uparrow$ \\(x$10^{-2}$)} \\
    \hline

    %%%%%%%%%%%% RTOG %%%%%%%%%%%%
    \hline
    \multirow{6}{*}{\parbox{2.0cm}{\centering ID \\ ------------ \\ H\&N CT \\ (RTOG)}} 
     & Det        & 84.2 ± 2.7 & 9.0 ± 2.1 & 73.0 ± 5.7 & 21.0 ± 4.8\\
     & Ensemble   & \underline{85.0 ± 2.6} & 8.6 ± 2.1 & \underline{78.6 ± 4.7} & \underline{25.7 ± 6.8}\\
     \cdashline{2-6}
     & Bayes      & 83.9 ± 2.6 & 8.6 ± 2.1 & 74.1 ± 5.4 & 22.1 ± 3.5\\
     & Bayes+AvU  & 83.6 ± 2.5 & \textbf{\underline{7.6 ± 2.5}}$^{\dag}$ & \textbf{76.1 ± 5.6}$^{\dag}$ & \textbf{25.1 ± 5.3}$^{\dag}$\\
     \cdashline{2-6}
     & BayesH     & 83.6 ± 2.9 & 9.2 ± 2.6 & 70.4 ± 7.0 & 20.1 ± 3.8\\
     & BayesH+AvU & \textbf{84.1 ± 2.7} & \textbf{8.4 ± 2.4}$^{\dag}$ & \textbf{74.1 ± 5.4}$^{\dag}$ & \textbf{21.3 ± 4.6}$^{\dag}$\\

    %%%%%%%%%%%% STRSeg %%%%%%%%%%%%
    \hline
    \multirow{6}{*}{\parbox{2.0cm}{\centering OOD \\ ------------ \\ H\&N CT \\ (STRSeg)}} 
     & Det       & 78.1 ± 4.6 & 12.9 ± 2.6  & 62.2 ± 4.5 & 24.1 ± 3.7\\
     & Ensemble  & \underline{78.6 ± 5.2} & \underline{10.6 ± 2.4}  & 64.7 ± 4.9 & 28.2 ± 5.1\\
     \cdashline{2-6}
     & Bayes     & 75.0 ± 9.9 & 12.4 ± 4.0  & 64.8 ± 5.0 & 27.7 ± 5.8\\
     & Bayes+AvU & \textbf{76.3 ± 7.6}$^{\dag}$ & \textbf{12.1 ± 3.7}  & \textbf{\underline{65.8 ± 5.0}}$^{\dag}$ & \textbf{\underline{30.1 ± 6.5}}$^{\dag}$\\
    \cdashline{2-6}
     & BayesH     & 77.5 ± 6.6 & 12.6 ± 3.3 & 61.1 ± 4.1 & 23.5 ± 4.7\\
     & BayesH+AvU & \textbf{78.8 ± 5.1}$^{\dag}$ & \textbf{12.1 ± 3.2} & \textbf{64.8 ± 3.8}$^{\dag}$ & \textbf{23.8 ± 4.0}$^{\dag}$\\

    %%%%%%%%%%%% ProsMedDec %%%%%%%%%%%%
    \hline
    \multirow{6}{*}{\parbox{2.0cm}{\centering ID \\ ------------ \\ Prostate \\ MR \\ (PrMedDec)}} 
     & Det        & 84.1 ± 5.6 & 12.9 ± 6.0 & 92.5 ± 5.7 & 28.0 ± 3.7\\
     & Ensemble   & 84.5 ± 5.7 & 11.3 ± 6.5 & 94.3 ± 4.3 & 30.0 ± 4.6\\
     \cdashline{2-6}
     & Bayes      & 84.0 ± 5.8 & \underline{8.6 ± 4.7}  & 94.7 ± 3.1 & 29.1 ± 4.8\\
     & Bayes+AvU  & \textbf{\underline{84.9 ± 6.9}} & 8.9 ± 6.0  & \textbf{\underline{95.7 ± 3.2}}$^{\dag}$ & \textbf{\underline{30.5 ± 4.5}}$^{\dag}$\\
     \cdashline{2-6}
     & BayesH     & 82.3 ± 5.2 & 9.3 ± 4.3  & 93.6 ± 2.9 & 28.4 ± 4.2\\
     & BayesH+AvU & \textbf{84.5 ± 6.3}$^{\dag}$ & 9.4 ± 6.5  & \textbf{94.9 ± 3.1}$^{\dag}$ & \textbf{30.1 ± 4.9}$^{\dag}$\\

    %%%%%%%%%%%% PR12 %%%%%%%%%%%%
    \hline
    \multirow{6}{*}{\parbox{2.0cm}{\centering OOD \\ ------------ \\ Prostate \\ MR \\ (PR12)}} 
     & Det        & 74.2 ± 12.6 & 15.6 ± 6.3 & 87.9 ± 7.5 & 22.1 ± 6.2\\
     & Ensemble   & \underline{76.3 ± 12.2} & \underline{9.7 ± 5.0}  & \underline{91.6 ± 5.2} & \underline{28.4 ± 5.7}\\
     \cdashline{2-6}
     & Bayes      & 70.6 ± 16.6 & 11.8 ± 7.2 & 89.1 ± 7.4 & 25.7 ± 5.1\\
     & Bayes+AvU  & \textbf{76.3 ± 12.6}$^{\dag}$ & \textbf{11.4 ± 6.7}$^{\dag}$ & \textbf{90.6 ± 6.9}$^{\dag}$ & \textbf{26.2 ± 7.4}$^{\dag}$\\
    \cdashline{2-6}
     & BayesH     & 71.3 ± 14.4 & 12.1 ± 6.7 & 88.9 ± 6.3 & 25.1 ± 4.9\\
     & BayesH+AvU & \textbf{74.1 ± 13.8}$^{\dag}$ & \textbf{11.9 ± 6.2}$^{\dag}$ & \textbf{89.9 ± 6.2}$^{\dag}$ & \textbf{25.9 ± 5.4}$^{\dag}$\\
    
    \hline
    \end{tabular}
    
    \label{tab:results_table_vsm_bayesh}
\end{table}

\end{document}